\documentclass[preprint,12pt,authoryear]{elsarticle}

\usepackage{fullpage}
\usepackage[utf8]{inputenc}
\usepackage[T1]{fontenc}
\usepackage{amsthm}
\usepackage{amsmath,amssymb,amsfonts}
\usepackage{bm}
\usepackage{mathtools}
\usepackage{booktabs}
\usepackage{array}
\usepackage{microtype}
\usepackage{xcolor}
\usepackage[shortlabels]{enumitem}
\usepackage[linesnumbered,algoruled,lined]{algorithm2e}
\SetAlFnt{\footnotesize}
\SetAlgoNlRelativeSize{-1}
\usepackage{graphicx}
\usepackage{tikz}
\usetikzlibrary{arrows,arrows.meta,shapes,decorations}
\usepackage{pgfplots}
\pgfplotsset{compat=newest}
\usepackage{chngcntr}
\usepackage[hidelinks]{hyperref}

\journal{International Journal of Approximate Reasoning}
\biboptions{sort&compress}

\newcommand{\setB}{\mathcal{B}}
\newcommand{\setC}{\mathcal{C}}

\newcommand{\setG}{\mathcal{G}}
\newcommand{\setH}{\mathcal{H}}
\newcommand{\setI}{\mathcal{I}}
\newcommand{\setJ}{\mathcal{J}}

\newcommand{\setM}{\mathcal{M}}
\newcommand{\setN}{\mathcal{N}}

\newcommand{\setP}{\mathcal{P}}

\newcommand{\setV}{\mathcal{V}}

\newcommand{\setX}{\mathcal{X}}
\newcommand{\setY}{\mathcal{Y}}
\newcommand{\setZ}{\mathcal{Z}}

\newcommand{\bw}{\mathbf{w}}
\newcommand{\bn}{\mathbf{n}}
\newcommand{\bmm}{\mathbf{m}}

\newcommand{\bB}{\mathbf{B}}
\newcommand{\bC}{\mathbf{C}}
\newcommand{\bD}{\mathbf{D}}

\newcommand{\bZ}{\mathbf{0}}
\newcommand{\bW}{\mathbf{W}}

\newcommand{\bI}{\mathbf{I}}

\newcommand{\reals}{\mathbb{R}}
\newcommand{\pml}{p_{ml}}

\newcommand{\uleaf}{mleaf }

\tikzstyle{observed}=[draw, circle, white, text=black, minimum size=8mm]
\tikzstyle{latent}=[draw, circle, blue!75, text=black, minimum size=8mm]
\tikzstyle{deterministic}=[draw, circle, double, blue!75, text=black, minimum size=8mm]
\tikzstyle{noise}=[draw, text=black]

\tikzset{causal/.style={-{Triangle[length=3.5pt, width=4pt]}, line width=1pt},
         exogenous/.style={-{Triangle[length=2.5pt, width=3pt]}, line width=1pt, blue!75},
         difference/.style={-{Triangle[length=2.5pt, width=3pt]}, line width=1pt, red},
         approved/.style={-{Triangle[length=2.5pt, width=3pt]}, line width=1pt, green!90!black}
}

\tikzstyle{myshape}=[
  thick,draw=white,fill=green!25,rounded corners=0.01]

\tikzstyle{aog}=[
  thick,draw=green,dashed, line width=1.5pt, dash pattern=on 6pt off 6pt, rounded corners=0.01]

\newcommand{\vki}{V_{k_i}}
\newcommand{\vkl}{V_{k_l}}
\newcommand{\wzj}{\bW_{K\cup \{Z_j\}}^J}
\newcommand{\wki}{\bW_{K\cup \{\vki\}}^J}
\newcommand{\wkl}{\bW_{K\cup \{\vkl\}}^J}

\newcommand{\ww}[1]{Rank(\bW_{K\cup \{#1\}}^J)}

\theoremstyle{definition}
\newtheorem{definition}{Definition}
\newtheorem{condition}{Condition}
\newtheorem{assumption}{Assumption}
\theoremstyle{plain}
\newtheorem{theorem}{Theorem}

\newtheorem{proposition}{Proposition}
\theoremstyle{remark}
\newtheorem{remark}{Remark}
\newtheorem{example}{Example}
\DeclareMathOperator{\Rank}{Rank}
\renewcommand{\ww}[1]{\Rank(\bW_{K\cup \{#1\}}^J)}

\begin{document}

\begin{frontmatter}

\title{Causal Discovery in Linear Models with Unobserved Variables and Measurement Error}

\author[inst1]{Yuqin Yang}
\author[inst2]{Mohamed Nafea}
\author[inst3]{Negar Kiyavash}
\author[inst4,inst5]{~\\Kun Zhang}
\author[inst6]{AmirEmad Ghassami\corref{cor1}}
\cortext[cor1]{Corresponding author}
\ead{ghassami@bu.edu}

\address[inst1]{Georgia Institute of Technology, Atlanta, GA 30332, USA}
\address[inst2]{Missouri University of Science and Technology, Rolla, MO 65409, USA}
\address[inst3]{\'Ecole Polytechnique F\'ed\'erale de Lausanne, Lausanne, Switzerland}
\address[inst4]{Carnegie Mellon University, Pittsburgh, PA 15213, USA}
\address[inst5]{Mohamed bin Zayed University of Artificial Intelligence, Masdar City, Abu Dhabi, UAE}
\address[inst6]{Boston University, Boston, MA 02215, USA}

\begin{abstract}
The presence of unobserved common causes and measurement error poses two major obstacles to causal structure learning, since ignoring either source of complexity can induce spurious causal relations among variables of interest. We study causal structure learning in linear systems where both challenges may occur simultaneously. We introduce a causal model called LV-SEM-ME, which contains four types of variables: directly observed variables, variables that are not directly observed but are measured with error, the corresponding measurements, and variables that are neither observed nor measured. Under a separability condition---namely, identifiability of the mixing matrix associated with the exogenous noise terms of the observed variables---together with certain faithfulness assumptions, we characterize the extent of identifiability and the corresponding observational equivalence classes. We provide graphical characterizations of these equivalence classes and develop recovery algorithms that enumerate all models in the equivalence class of the ground truth. We also establish, via a four-node union model that subsumes instrumental variable, front-door, and negative-control-outcome settings, a form of identification robustness: the target effect remains identifiable in the broader LV-SEM-ME model even when the assumptions underlying the specialized identification formulas for the corresponding submodels need not all hold simultaneously.
\end{abstract}


\end{frontmatter}
\section{Introduction}
Causal structure learning, also known as causal discovery, from observational data has been studied extensively. Most existing work assumes that there are no unobserved common causes and that variables are measured without error. Under these assumptions, the underlying structure is identifiable up to Markov equivalence in general \citep{spirtes2000causation,chickering2002optimal}, and it can be fully identified under additional model assumptions such as linearity with non-Gaussian noise \citep{shimizu2006linear,shimizu2011directlingam}. In many real-world settings, however, researchers cannot observe all relevant variables and therefore cannot rule out latent confounding or measurement error. This motivates causal discovery methods that can accommodate both challenges.

These two challenges have each been studied extensively, yet disjointly. For unobserved common causes, constraint-based algorithms such as FCI are widely used \citep{spirtes2000causation}, but they often leave many edge directions unresolved. In linear models, methods that exploit non-Gaussianity also generally fail to identify the structure uniquely in the presence of latent confounding. Approaches such as latent variable LiNGAM \citep{hoyer2008estimation,salehkaleybar2020learning} and partially observed LiNGAM \citep{adams2021identification} provide graphical conditions for unique identification, but these conditions are often nontrivial.

Fewer works study causal discovery under measurement error. Most existing work \citep{Silva06,kummerfeld2016causal,Xie20,pmlr-v162-xie22a} assumes that each unobserved (but measured) variable has at least two measurements. This assumption often enables unique identification, but it may be unrealistic in many applications.\footnote{As discussed in Section \ref{sec:desc}, we do not impose this assumption.} Without it, \citet{halpern2015anchored} consider binary measured variables, \citet{saeed2020anchored} study nonlinear relations among measured variables with Gaussian measurement error, and \citet{zhang2018causal} provide sufficient conditions for linear Gaussian and non-Gaussian models. However, this literature does not characterize observational equivalence when the model is not uniquely identifiable.

In this work, we study the problem of causal discovery from observational data in the presence of both aforementioned challenges, i.e., in settings with both unobserved common causes and measurement error. We consider a special type of linear structural equation model (SEM) as the underlying data generating process which includes four types of variables: variables that are directly observed (called \emph{observed variables}), variables that are not directly observed but are measured with error (called \emph{measured variables}), the corresponding \emph{measurement variables}, and variables that are neither directly observed nor measured (called \emph{unobserved variables}). We refer to this model as \emph{linear latent variable SEM with measurement error} (linear LV-SEM-ME). We study the identifiability of linear LV-SEM-MEs in a setup where the independent exogenous noise terms
that causally (directly or indirectly) affect each observed variable can be distinguished from each other. That is, the mixing matrix of the linear system that transforms exogenous noise terms to observed variables is recoverable up to permutation and scaling of the columns. This can be satisfied, for example, if all independent exogenous noise terms are non-Gaussian.
We note that measurement error challenge is essentially a special case of unobserved variable challenge, in which we observe a measurement of an underlying unobserved variable of interest. Yet, the observed measurement variable usually has special properties (such as not being affected by other variables) that can be leveraged to improve the identification power. Hence, our point of view in this work is to allow for coexistence of the challenges of unobserved common causes and measurement error, while leveraging the properties of the measurement variables to improve identification.

We study identifiability of linear LV-SEM-ME under two faithfulness assumptions. The first, which we call \emph{conventional faithfulness}, excludes zero total causal effects from a variable to its descendants. The second, which we call \emph{LV-SEM-ME faithfulness}, further excludes certain parameter cancellations and proportionalities. Both assumptions fail only on measure-zero subsets of the parameter space. Under conventional faithfulness, we show that the model is identifiable up to an equivalence class characterized by an ordered grouping of variables, which we call the \emph{ancestral ordered grouping} (AOG). Under LV-SEM-ME faithfulness, the model is identifiable up to a finer equivalence class characterized by the \emph{direct ordered grouping} (DOG). 
We provide a graphical characterization of the elements of the equivalence class in which the induced graph on each ordered group includes a star structure where any member (variable) of the group is a potential center of the star. Specifically, every element in the equivalence class corresponds to distinct assignments of the centers of the star graphs, yet it possesses the same ordered grouping of variables and the same unlabelled structure on each group as the rest of the elements in the equivalence class.
Models in the same AOG equivalence class are consistent with the same set of causal orders among groups, and models in the same DOG equivalence class share the same \emph{unlabeled graph structure}, i.e., the causal diagrams are isomorphic. Lastly, we provide a recovery algorithm that returns all models in the AOG and DOG equivalence classes. We further show in Section~\ref{app:application} that our proposed framework yields a concrete form of \emph{identification robustness}: in a four-node union model that subsumes instrumental variable, front-door, and negative-control-outcome settings, the target effect remains identifiable under the broader LV-SEM-ME formulation even when the assumptions behind those three specialized identification formulas may fail simultaneously. Thus, one need not know \emph{a priori} which of the three classical designs is correct in order to identify the effect within this broader model family. 

Preliminary versions of some of the ideas of this paper appeared in \citep{yang2022causal2}. That work, however, treated confounding and measurement error separately and did not address settings in which these two challenges coexist. It also adopted different definitions of AOG and DOG; see Remark~\ref{rmk:diff}. In addition, unlike \citep{yang2022causal2}, the present paper develops the identification robustness perspective explicitly, clarifies the relationship between the proposed framework and instrumental variable, negative-control-outcome, and front-door models, and complements the theory with numerical experiments that compare our general recovery procedure against the corresponding specialized estimators and assess the sensitivity of DOG recovery to perturbations in the mixing matrix.

The rest of the paper is organized as follows. In Section~\ref{sec:desc}, we provide a formal description of the model and the problem. In Section~\ref{sec:id}, we present the identifiability analysis under our two faithfulness assumptions. We first establish the corresponding results for two submodels, one involving only unobserved common causes and the other involving only measurement error, and then consider the general case in which these two challenges coexist. In Section~\ref{sec:alg}, we present our recovery algorithms for the LV-SEM-ME. In Section~\ref{app:application}, we use the framework to unify instrumental variable, negative-control-outcome, and front-door models and to highlight an identification robustness result. In Section~6, we report two sets of numerical experiments: one compares our general recovery procedure with specialized estimators, and the other studies the robustness of DOG recovery to noisy and finite-sample estimates of the mixing matrix. We conclude in Section~\ref{sec:conc}.

\section{Model Description}
\label{sec:desc}
\paragraph{Notations} We use upper-case letters with subscript for variables, upper-case letters without subscript for vectors and bold upper-case letters for matrices. For two vectors or variables $X,Y$, we use $[X, Y]$ to represent the horizontal concatenation of $X$ and $Y$, and $[X;Y]$ to represent the vertical concatenation of $X$ and $Y$. 

We start with a formal definition of the model that we consider in this work.
\begin{definition}[General linear LV-SEM-ME]
A general linear LV-SEM-ME consists of two sets of variables $\setV$ and $\setX$. Variables in $\setV$ can be arranged in a causal order, and each variable $V_i\in \setV$ is generated as a linear combination of a subset $Pa(V_i)\subset \setV$ (called its direct parents), plus an exogenous noise term $N_{V_i}$, where $\{N_{V_i}\}_{V_i\in \setV}$ are jointly independent. Further, $\setV$ can be partitioned into three sets $\setY$, $\setZ$ and $\setH$. Variables in $\setY$ are observed (without error). Variables in $\setZ$ are measured with error, where each variable $X_i\in \setX$ is a noisy measurement of one corresponding variable $Z_i\in \setZ$ plus an exogenous noise term $N_{X_i}$ (which we call measurement error of $Z_i$). Variables in $\setH$ are neither observed nor measured with error. We refer to variables in $\setH$, $\setY$, $\setZ$, and $\setX$ as unobserved variables, observed variables, measured variables, and measurements, respectively.
\end{definition}

We define a measured leaf variable (mleaf variable) as a measured variable in $\setZ$ that has no other children besides its noisy measurement. We define a \emph{cogent variable} as a variable in $\setZ\cup\setY$ that is not an mleaf. 
As mentioned earlier, we study the problem of recovering the linear LV-SEM-ME from observations of $(\setY, \setX)$. For identifiability from observational data, we impose the following two restrictions on the model. 
\begin{itemize}
    \item Firstly, as discussed in \citep{zhang2018causal, yang2022causal2}, for any mleaf variable $Z_i$, the exogenous noise term $N_{Z_i}$ is not distinguishable from its measurement error $N_{X_i}$. Specifically, for any two models that only differ in $Z_i$ and $X_i$ for some mleaf variable $Z_i$ but have the same sum $N_{Z_i}+N_{X_i}$, they have the same observational distribution. This follows because $Z_i$ is not observed, and $N_{Z_i}$ only influences its noisy measurement $X_i$ and no other observed variables in $\setY$. Therefore, for purposes of identifiability, we work with the equivalent representation in which $N_{Z_i}=0$ for all mleaf variables; that is, mleaf variables are deterministically generated from their direct parents.
    \item Secondly, we assume that variables in $\setH$ are all root variables (i.e., have no direct parents) and confounders (i.e., have at least two children). This is because for any linear latent variable model with a non-root latent variable, there exists an equivalent latent variable model where the latent variables are all roots, such that both models have the same joint distribution across observed variables and total causal effect between any pair of observed variables \citep{hoyer2008estimation}. 
\end{itemize}

Due to the aforementioned restrictions, we focus on recovering the subset of linear LV-SEM-ME, called \emph{canonical LV-SEM-ME}, defined as follows.

\begin{definition}[Canonical linear LV-SEM-ME]
\label{def:canonical}
A canonical LV-SEM-ME is an LV-SEM-ME where (i) variables in $\setH$ are roots and confounders, and (ii) mleaf variables do not have distinct exogenous noise terms. 
\end{definition}

The matrix form of the canonical LV-SEM-ME can be written as
\begin{align}
H~~&=N_H,
\tag{1a}
\label{eq:1a}
\\
\begin{bmatrix}
Z^L\\
Z^{C}\\
Y
\end{bmatrix}
&=\bB H +
\begin{bmatrix}
\bD \\
\bC_Z \\
\bC_Y
\end{bmatrix}
\begin{bmatrix}
Z^{C}\\
Y
\end{bmatrix}
+ 
\begin{bmatrix}
\bm{0} \\
N_{Z^{C}}\\
N_Y
\end{bmatrix}
,
\tag{1b}
\label{eq:1b}
\\
X
~~&= 
\begin{bmatrix}
Z^L\\
Z^{C}
\end{bmatrix}
+ 
\begin{bmatrix}
N_{X^L}\\
N_{X^{C}}
\end{bmatrix}
,
\tag{1c}
\label{eq:1c}
\end{align}
where $H$, $Y$, and $X$ denote the vectors of unobserved variables, observed variables, and measurements, respectively. $Z^L$ denotes the vector of mleaf variables, and $Z^{C}$ denotes the vector of measured variables that are not mleaf variables. $N_{H}$, $N_{Y}$, and $N_{Z^{C}}$ are the corresponding exogenous-noise vectors. $N_{X^L}$ (resp. $N_{X^{C}}$) denotes the measurement error of the variables in $Z^L$ (resp. $Z^{C}$). 
Let the numbers of variables in $H$, $Y$, $Z^{C}$, and $Z^L$ be $p_H$, $p_Y$, $p_{Z^{C}}$, and $p_{ml}$, respectively. The number of cogent variables is $p_{c}= |Y| + |Z^{C}|=p_Y + p_{Z^{C}}$, and the total number of observed variables is $p=|Y|+|X|=p_Y + p_{Z^{C}} + p_{ml}$. 
$\bB\in \reals^{p\times p_{H}}$ encodes the causal connections from the unobserved variables $H$ to $[Z^L;Z^C;Y]$, $\bC\in \reals^{p_c \times p_c}$ encodes the causal connections among cogent variables and is partitioned into $[\bC_Z; \bC_Y]$ according to $[Z^{C}; Y]$, and $\bD\in \reals^{\pml \times p_c}$ encodes the causal connections from cogent variables to mleaf variables. Note that the right-hand side of Equation \eqref{eq:1b} does not depend on $Z^L$ because variables on the left-hand side cannot have mleaf variables as parents.

We define the causal diagram of a linear LV-SEM-ME as a directed graph, where the nodes are all variables in $\setV\cup \setX$. For any two variables $W_1, W_2\in \setV\cup \setX$, there is a directed edge from $W_1$ to $W_2$ if and only if $W_1\in Pa(W_2)$. Due to the causal order of $ \setV\cup \setX$, the causal diagram is acyclic. 

\begin{figure}[t]
    \centering
\begin{tikzpicture}[very thick, scale=0.5]
\foreach \place/\name in {{(-4,-1.25)/X_1}, {(0,-3)/X_2}, {(4,-1.25)/Y_3}}
    \node[observed, label=center:{\Large $\name$}] (\name) at \place {};

\foreach \place/\name in {{(-2,2.5)/Z_1}, {(0,0)/Z_2}, {(2,2.5)/H}}
    \node[latent, black, label=center:{\Large $\name$}] (\name) at \place {};

\foreach \source/\dest in {Z_1/X_1, Z_1/Z_2, H/Z_2, H/Y_3, Z_2/X_2}
    \path[causal] (\source) edge (\dest);

\end{tikzpicture}
\hspace{10mm}
\begin{tikzpicture}[very thick, scale=0.5]

\foreach \place/\name in {{(-2,2.5)/Z_1}, {(4,0)/Y_3}}
    \node[observed, label=center:{\Large $\name$}] (\name) at \place {};

\foreach \place/\name in {{(0,0)/Z_2}}
    \node[deterministic, label=center:{\Large $\name$}] (\name) at \place {};

\foreach \place/\name in {{(2,2.5)/H}}
    \node[latent, label=center:{\Large $\name$}] (\name) at \place {};

\foreach \source/\dest in {Z_1/Z_2, H/Z_2, H/Y_3}
    \path[causal] (\source) edge (\dest);

\node[] () at (0,-3) {};

\end{tikzpicture}
\caption{\textbf{Left}: Diagram of the model in Example \ref{example:ex1}. Black circles represent variables that are not observed (in the noncanonical form). \textbf{Right}: Diagram of the canonical model. Double circles represent mleaf variables in $\setZ^L$, and blue circles represent unobserved variables in $\setH$.}
    \label{fig:figure_ex1}
\end{figure}
\begin{example} \label{example:ex1}
    Figure \ref{fig:figure_ex1} shows an example of a causal diagram including unobserved variable $H$, observed variable $\{Y_3\}$, measured variables $\{Z_1,Z_2\}$ and their corresponding measurements $\{X_1, X_2\}$. The generating model is as follows.
    \begin{equation*}
        \begin{aligned}
            H &= N_{H},\\
            Z_1 &= N_{Z_1},\\
            Z_2 &= b_2 H + a_{21} Z_1 + N_{Z_2},\\
            Y_3 &= b_3 H + N_{Y_3},\\
            X_1 &= Z_1 + N_{X_1},\\
            X_2 &= Z_2 + N_{X_2}.
        \end{aligned}
    \end{equation*}
We note that $Z_2$ is an mleaf variable; it has no other children except for $X_2$. $\{Z_1, Y_3\}$ are cogent variables. Therefore, in the canonical form, the exogenous noise of $Z_2$ is $0$, and the measurement error of $Z_2$ is $N_{Z_2} + N_{X_2}$. In the matrix form of Equation \eqref{eq:1b}, we have $Z^L=[Z_2]$, $Z^C=[Z_1]$, $Y=[Y_3]$, $\bB = [b_2, 0, b_3]^{\top}$, $\bD = [a_{21}, 0]$, and $\bC_Z = \bC_Y= [0, 0]$.

\end{example}

\paragraph{Problem description} 
We consider a setting with known \emph{observability indicators}, that is we know whether each variable is observed without error (i.e., belongs to $Y$) or measured with error (i.e., belongs to $X$).
Suppose we have $n$ i.i.d. observations of the variables $\{X,Y\}$.
The task is to recover all linear LV-SEM-MEs which have the same observational distribution up to the noise distributions.

We first consider the problem for two special submodels in Section \ref{sec:ME-LV-sep}: 
1) If $\setH=\emptyset$, i.e., all unobserved variables have noisy measurements, then the model is linear SEM-ME.
2) If $\setZ=\emptyset$, i.e., all unobserved variables are roots and confounders, then the model is linear LV-SEM. Identification analysis for these two special cases were also studied in \citep{yang2022causal2}, yet different techniques were used in that work. 
We then study the general form in Section \ref{sec:ME-LV-joint}, where both challenges can be present in the system simultaneously.

\section{Identification Analysis}
\label{sec:id}
In this section we study identification for our model of interest. We start by looking at the LV-SEM and SEM-ME separately in Subsection \ref{sec:ME-LV-sep}; we consider identification in the presence of both unobserved variables and measured variables in Subsection \ref{sec:ME-LV-joint}, which is our main identification result.
In both subsections, we study identification under two faithfulness assumptions where, as will be discussed shortly, the first one, referred to as the conventional faithfulness, is a weaker assumption.

\subsection{Identifiability of SEM-ME and LV-SEM}
\label{sec:ME-LV-sep}
\subsubsection{Identification Assumptions}\label{sec:assumptions}
We first present two assumptions for identifiability of SEM-ME and LV-SEM: a \emph{separability assumption} and a \emph{faithfulness assumption}. For LV-SEM-ME we will require one additional assumption, introduced in Section \ref{sec:ME-LV-joint}.

\paragraph{Separability assumption} 

We first deduce the mixing matrix that transforms independent exogenous noise terms to the observed variables $[X; ~Y]$. To simplify the deduction, we rewrite Equation \eqref{eq:1b} as follows, by considering cogent variables ($Z^C$ and $Y$) as a single vector:
\setcounter{equation}{1}
\begin{equation}
\begin{bmatrix}
Z^L\\
V^{C}
\end{bmatrix}
=
\begin{bmatrix}
\bB^{L}\\
\bB^{C}
\end{bmatrix}
H +
\begin{bmatrix}
\bD \\
\bC
\end{bmatrix}
V^C
+ 
\begin{bmatrix}
\bm{0} \\
N_{V^{C}}
\end{bmatrix}
,
\label{eq:2b}
\end{equation}
where $V^{C}=[Z^C; Y]$ denotes the vector of cogent variables, and $N_{V^{C}}=[N_{Z^{C}}; N_Y]$ denotes the corresponding vector of exogenous noises. 
$\bB$ is partitioned into $[\bB^L;\bB^C]$ according to $[Z^L;V^C]$. 
From Equations \eqref{eq:1a} and \eqref{eq:2b}, we can write variables in $[Z^L; V^C]$ as linear combinations of the exogenous noise terms:

\begin{align} 
    \begin{bmatrix}
    Z^{L} \\
    V^{C} \\
    \end{bmatrix} =
    \bW^*
    \begin{bmatrix}
    N_{H} \\
    N_{V^{C}}
    \end{bmatrix},
    \quad \text{where} ~ \bW^* = 
    \begin{bmatrix}
    &\bB^L + \bD(\bI-\bC)^{-1}\bB^{C} &\bD(\bI-\bC)^{-1} \\
    &(\bI-\bC)^{-1}\bB^{C} &(\bI-\bC)^{-1} 
    \end{bmatrix},
    \label{eq:Wstar}
\end{align}
and $\bI$ represents the $p_c\times p_c$ identity matrix. Lastly, combined with Equation \eqref{eq:1c}, the overall mixing matrix $\bW$ can be written as
\begin{align} 
    \begin{bmatrix}
    X \\
    Y
    \end{bmatrix}
    =
    \underbrace{
    \begin{bmatrix}~
    \bW^* & 
    \begin{matrix}
    \bI \\
    \bZ
    \end{matrix}
    ~
    \end{bmatrix}
    }_\bW
    \begin{bmatrix}
    N_{H} \\
    N_{V^{C}} \\
    N_{X^{L}} \\
    N_{X^{C}} 
    \end{bmatrix}
    \label{eq:W}
    .
\end{align}
We note that because mleaf variables have no exogenous noise, each column in $\bW^*$ either has at least two non-zero entries, or has one non-zero entry where the non-zero entry is not in $X$. 

\setcounter{example}{0}
\begin{example}[Continued]
The mixing matrix $\bW$ in Example \ref{example:ex1} can be written as
\begin{equation*}
\begin{bmatrix}
    X_2 \\
    X_1 \\
    Y_3
    \end{bmatrix}
    =
    \begin{bmatrix}~
    b_2 & a_{21} & 0 & 1 & 0 \\
    0 & 1 & 0 & 0 & 1 \\
    b_3 & 0 & 1 & 0 & 0 \\
    \end{bmatrix}
    \begin{bmatrix}
    N_{H} \\
    N_{Z_1} \\
    N_{Y_3} \\
    N_{X_2} \\
    N_{X_1}
    \end{bmatrix}.
\end{equation*}
The leftmost three columns correspond to $\bW^*$, and $\bI$ is of dimension $2\times 2$.
\end{example}
We are now ready to state our requirement regarding recoverability of the mixing matrix.
\begin{assumption}[Separability] \label{assumption:separability}
The mixing matrix $\bW$ in Equation \eqref{eq:W} can be recovered from observations of $[X;\; Y]$ up to permutation and scaling of its columns.
\end{assumption}

We call a linear LV-SEM-ME separable if the corresponding mixing matrix satisfies Assumption \ref{assumption:separability}.
The separability assumption states that the independent exogenous-noise terms in the mixture of Equation \eqref{eq:W} can be separated, meaning that the mixing matrix can be recovered up to permutation and scaling of its columns. An example of a setting where this assumption holds is when all exogenous noises are non-Gaussian. In this case, if the model satisfies the requirement in \citep[Theorem 1]{eriksson2004identifiability}, then overcomplete Independent Component Analysis (ICA) can be used to recover the mixing matrix up to permutation and scaling of its columns. Another setting in which separability holds is when the noise terms are piecewise-constant functionals satisfying mild conditions \citep{behr2018multiscale}. By contrast, if all exogenous-noise terms are Gaussian, then the mixing matrix can in general be recovered only up to an orthogonal transformation.

Under separability, the matrix $\bW^*$ is also recoverable up to permutation and scaling of its columns. Suppose the recovered overall mixing matrix is $\hat{\bW}$, and suppose we also know for each observed variable whether it belongs to $\setY$ or $\setX$, that is, whether each variable is directly observed or measured with error. Then $\bW^*$ can be obtained by removing from $\hat{\bW}$ the one-hot columns whose non-zero entry appears in a row corresponding to a variable in $\setX$.
The justification of this approach is as follows: If a variable $X_i$ is measured with error, then there must exist one column in $\hat{\bW}$ that corresponds to the measurement error $N_{X_i}$ (or $N_{X_i} + N_{Z_i}$ for mleaf variables in the original form). This column has only one non-zero entry in the row corresponding to the measurement $X_i$. The columns that we remove in the procedure above correspond to such measurement errors. We note that by removing these columns from $\hat{\bW}$ we are not losing any information. 
This is due to the fact that we can simply recover a matrix (permutationally) equivalent to $\hat{\bW}$ from $\bW^*$:
Since we know which variables are measured with error, we can simply add the corresponding one-hot column vectors back to the matrix. The order of adding these is irrelevant (no information) since it is arbitrary in the original $\hat{\bW}$ as well (recall that $\bW$ is identifiable up to permutation and scaling of the columns).

\paragraph{Faithfulness assumption}

For each variable $V_i\in \setV$, let $An(V_i)$ denote the set of ancestors of $V_i$ (excluding $V_i$ itself), and let $An_V(V_i)=An(V_i)\setminus \setH$ denote the subset of cogent ancestors. Define the possible parent set of $V_i$, denoted $PP(V_i)$, as the union of $An_V(V_i)$ and the set of mleaf variables whose parent sets are subsets of $An(V_i)$ (excluding $V_i$ itself when $V_i$ is an mleaf).
For two sets of variables $J\subseteq \setH\cup \setV^C$ and $K\subseteq \setZ\cup\setY$, let $\bW_{K}^{J}$ denote the submatrix of $\bW^*$ whose rows correspond to the variables in $K$ and whose columns correspond to the exogenous-noise terms of the variables in $J$. A set of variables $\setB$ is called a bottleneck from $J$ to $K$ if every directed path from a variable in $J$ to a variable in $K$ contains at least one variable in $\setB$ (possibly the start or end node). It is called a minimal bottleneck if no other bottleneck from $J$ to $K$ has fewer variables.

We now state the two versions of faithfulness used in our identification results.

\begin{assumption}[Conventional faithfulness]\label{assumption:conv_Faithfulness}
    The total causal effect of any variable $V_i\in\setV$ on its descendant $V_j\in\setV$ is not zero. 
\end{assumption}

\begin{assumption}[LV-SEM-ME faithfulness]\label{assumption:lv_me_faithfulness}
For each variable $V_i\in \setZ\cup\setY$ and any pair of subsets $(J,K)$, $J\subseteq \setH\cup\setV^{C}$ and $K\subseteq \setZ\cup\setY$ that satisfies at least one of the conditions below, the rank of the submatrix $\bW_{K\cup \{V_i\}}^J$ is equal to the size of minimal bottleneck from $J$ to $K\cup \{V_i\}$:
\begin{enumerate}[(a)]
    \item $J\subseteq An(V_i)$, $K\subseteq PP(V_i)$;
    \item $J\subseteq An(V_i) \setminus \{V_j\}$, $K\subseteq PP(V_j)$, when $V_i$ is a mleaf variable and $V_j$ is a parent of $V_i$.
\end{enumerate}
\end{assumption}

Assumption \ref{assumption:conv_Faithfulness} is standard in the literature, and we therefore refer to it as conventional faithfulness. It requires that when multiple causal paths exist from any (observed or unobserved) variable to its descendants, their combined effect (i.e.,
sum of products of path coefficients) is not equal to zero.
Note that Assumption \ref{assumption:conv_Faithfulness} is a special case of Assumption \ref{assumption:lv_me_faithfulness}(a) with $K=\emptyset$ and $J$ being a singleton set consisting of any ancestor of $V_i$.
The intuition of Assumption \ref{assumption:lv_me_faithfulness} is as follows. 
The structure of the causal diagram in the data generating process implies proportionality in the corresponding entries in the mixing matrix. For example, in the structure $V_1\to V_2\to V_3$, the mixing matrix with rows corresponding to $\{V_2, V_3\}$ and columns corresponding to $\{V_1, V_2\}$ is of rank 1.
However, there may exist extra proportionality among the entries in the mixing matrix that is not enforced by the graph. 
This extra proportionality may result in the data distribution corresponding to an alternative model that does not always happen.
The faithfulness assumption rules out such extra proportionality in the generating model. A related bottleneck-faithfulness condition was proposed by \citet{adams2021identification}, who consider arbitrary pairs of subsets $(J,K)$. Assumption \ref{assumption:lv_me_faithfulness} is strictly weaker than that condition.

\begin{remark}
\label{remark:faith}
Both Assumptions \ref{assumption:conv_Faithfulness} and \ref{assumption:lv_me_faithfulness} are violated with probability zero if all model coefficients are drawn randomly and independently from continuous distributions.
However, Assumption \ref{assumption:conv_Faithfulness} only concerns marginal independencies and rules out cancellations that would render an ancestor independent of its descendant.
In practice, due to sample size limitations, an approximate cancellation may be perceived as an actual cancellation. Therefore, although Assumptions \ref{assumption:conv_Faithfulness} and \ref{assumption:lv_me_faithfulness} both exclude only measure-zero parameter sets, Assumption \ref{assumption:conv_Faithfulness} may be easier to work with in finite samples. For this reason, we present separate results under Assumptions \ref{assumption:conv_Faithfulness} and \ref{assumption:lv_me_faithfulness}.
    
\end{remark}

\subsubsection{Identification Under Conventional Faithfulness}

We first present a graphical characterization of equivalence under conventional faithfulness, called Ancestral ordered grouping (AOG) equivalence, and then formally show that this notion of equivalence is the extent of identifiability in Theorem \ref{thm:aog}.

\begin{definition}[Ancestral ordered grouping (AOG)]
\label{def:ancestral_ordered_group_decomposition}
The AOG of a SEM-ME (resp. LV-SEM) is a partition of $\mathcal{Z}^L\cup\setV^C$ (resp. $\setH\cup\setV^C$) into distinct sets. This partition is described as follows:
\begin{enumerate}[(1)]
    \item Assign each cogent variable in $\setV^C$ to a distinct group.
    \item (i) \textbf{SEM-ME}: For each mleaf variable $Z_j\in \mathcal{Z}^L$, if it has one measured
    parent $Z_i\in\setZ\cap \setV^C$ such that $Z_j$ has no other parents, or all other parents of $Z_j$ are also ancestors of $Z_i$, assign $Z_j$ to the same group as $Z_i$. Otherwise, assign $Z_j$ to a separate group (with no cogent variable).

    (ii) \textbf{LV-SEM}: For each unobserved variable $H_j\in\setH$, if it has one cogent child $V_i\in\setV^C$ such that all other children of $H_j$ are also descendants of $V_i$, assign $H_j$ to the same group as $V_i$. Otherwise, assign $H_j$ to a separate group (with no cogent variable). 
\end{enumerate}
\end{definition}

\begin{definition}[AOG equivalence class]\label{def:AOG equivalence class}
The AOG equivalence class of a linear SEM-ME (resp. LV-SEM) is the set of models that have the same mixing matrix (up to permutation and scaling of its columns) and the same ancestral ordered groups.
\end{definition}

\paragraph{Graphical characterization} It was shown in \citep{yang2022causal2} that 
all models in the AOG equivalence class are consistent with the same set of causal orders among the groups (i.e., if a causal order on the groups is consistent with one model in the class, it is consistent with all the models in the class), but not necessarily all the same edges across the groups.\footnote{We note that the identification results in Theorem \ref{thm:aog} implies that AOG is the finest partition that satisfies this property under Assumption \ref{assumption:conv_Faithfulness}.\label{footnote} } That is, based on the AOG, the set of causal orders among groups is identifiable, but the edges across groups are not.
For a SEM-ME, according to Definition \ref{def:ancestral_ordered_group_decomposition}, there is at most one cogent variable in each ancestral ordered group. 
Furthermore, each observed cogent variable belongs to a separate group. Each
mleaf node $Z_j$ is assigned to the ancestral ordered group of at most one of its parents.
Hence, if a group has more than one variable, then there must be exactly one measured cogent variable, and the rest of the nodes are mleaf nodes which are children of this node. Thus the induced structure on each ancestral ordered group is a \emph{star graph}.
A similar property holds for LV-SEM, that is, if a group has more than one variable, then there must be exactly one cogent variable, and the rest of the variables are unobserved variables which are parents of this node. Define the center of the ancestral ordered group as the cogent variable, or the mleaf variable (resp. unobserved variable) if the group does not include a cogent variable. 
\citet{yang2022causal2} showed that fixing the center of each ancestral ordered group for SEM-ME, and fixing both the exogenous-noise term of the center and the scaling and permutation of the columns of $\bB$ for LV-SEM, leads to unique identification of the model.
Therefore, by considering all the candidates for the center, models in the same AOG equivalence class of a SEM-ME can be enumerated by switching the center of each group with other nodes that are in the same group. Models in the same AOG equivalence class of an LV-SEM can be enumerated by switching the exogenous noise of the center of each group with the noise of other nodes in the same group.

\begin{theorem} \label{thm:aog}
Under Assumptions \ref{assumption:separability} and \ref{assumption:conv_Faithfulness}, the linear SEM-ME (resp. LV-SEM) can be identified up to its AOG equivalence class.
\end{theorem}

\subsubsection{Identification Under LV-SEM-ME Faithfulness}

We first present a graphical characterization of equivalence under LV-SEM-ME faithfulness, called Direct ordered grouping (DOG) equivalence, and then formally show that this notion of equivalence is the extent of identifiability in Theorem \ref{thm:DOG}.

\begin{condition}[SEM-ME edge identifiability] \label{condition:me-id}
For a given edge from a measured cogent variable $Z_i$ to an mleaf variable $Z_l$, at least one of the following two conditions is satisfied:
(a) $Pa(Z_l)\setminus \{Z_i\}$ is not a subset of $Pa(Z_i)$. That is, there exists another parent $V_j$ of $Z_l$, which is not a parent of $Z_i$. 
(b) $Pa(Z_l)$ is not a subset of $\cap_{V_k\in Ch(Z_i)\setminus \{Z_l\}} Pa(V_k) $. That is, there exists a child $V_k$ of $Z_i$ and a parent $V_j$ of $Z_l$ such that $V_j$ is not a parent of $V_k$.
\end{condition}

\begin{condition}[LV-SEM edge identifiability]\label{condition:ur_id_new}
For a given edge from an unobserved variable $H_l$ to a cogent variable $V_i$, there exists another cogent child $V_j$ of $H_l$, such that at least one of the following two conditions is satisfied: 
(a) $V_i$ is not a direct parent of $V_j$.
(b) $Pa(V_i)$ is not a subset of $Pa(V_j)$. That is, there exists an observed (or unobserved) parent $V_k$ (or $H_k$) of $V_i$ that is not a parent of $V_j$.
\end{condition}

\begin{definition}[Direct ordered grouping (DOG)]
\label{def:ordered_group_decomposition}
The DOG of a linear SEM-ME (resp. LV-SEM) is a partition of $\mathcal{Z}^L\cup\setV^C$ (resp. $\setH\cup\setV^C$) into distinct sets. This partition is described as follows:
\begin{enumerate}[(1)]
\item 
Assign each cogent variable in $\setV^C$ to a distinct group.
\item (i) \textbf{SEM-ME}: For each mleaf variable $Z_j\in \mathcal{Z}^L$, if it has one measured parent $Z_i\in \setZ\cap\setV^C$ such that the edge from $Z_i$ to $Z_j$ violates Condition \ref{condition:me-id}, assign $Z_j$ to the same ordered group as $Z_i$. Otherwise, assign $Z_j$ to a separate ordered group (with no cogent variable).

(ii) \textbf{LV-SEM}: For each unobserved variable $H_j\in\setH$, if it has one cogent child $V_i\in \setV^C$ such that the edge from $H_j$ to $V_i$ violates Condition \ref{condition:ur_id_new}, assign $H_j$ to the same ordered group as $V_i$. Otherwise, assign $H_j$ to a separate ordered group (with no cogent variable).
\end{enumerate}

\end{definition}

\begin{definition}[DOG equivalence class]\label{def:DOG equivalence class}
The DOG equivalence class of a linear SEM-ME (resp. LV-SEM) is the set of models that have the same mixing matrix (up to permutation and scaling of its columns) and the same direct ordered groups.
\end{definition}

\paragraph{Graphical characterization}

It follows from Definition \ref{def:ordered_group_decomposition} that DOG is a refinement of AOG. Therefore, similar to AOG equivalence class, models in the DOG equivalence class are also consistent with the same set of causal orders among the groups,\footnote{Similar to the remark in Footnote \ref{footnote}, Theorem \ref{thm:DOG} implies that DOG is the finest partitioning that satisfies this property under Assumption \ref{assumption:lv_me_faithfulness}.} and
the induced structure on each direct ordered group is also a star graph. Further, models in the same DOG equivalence class of a SEM-ME and an LV-SEM can also be enumerated by switching the center of each group, and switching the exogenous noise of the center of each group with the noise of other nodes in the same group, respectively. For the properties that only hold for DOG
, it was shown in \citet{yang2022causal2} that models in the same DOG equivalence class have the same edges across the groups. Combined with the star structure within each group, the following proposition provides a graphical characterization of the DOG equivalence class for SEM-ME and LV-SEM. 

\begin{proposition}\label{propsition:lvsem_structure}
\begin{enumerate}[(a)]
    \item Models in the same DOG equivalence class of a SEM-ME have the same unlabeled graph structure, i.e., the causal diagrams of these models are isomorphic.
    \item Models in the same DOG equivalence class of an LV-SEM have the same graph structure.
\end{enumerate}

\end{proposition}

\begin{theorem}\label{thm:DOG}
Under Assumptions \ref{assumption:separability} and \ref{assumption:lv_me_faithfulness}, the linear SEM-ME (resp. LV-SEM) can be identified up to its DOG equivalence class with probability one.
\end{theorem}

The DOG equivalence class gives a substantially sharper characterization of the causal relations than the AOG equivalence class. This gain comes from strengthening Assumption \ref{assumption:conv_Faithfulness} to Assumption \ref{assumption:lv_me_faithfulness}. As emphasized in Remark \ref{remark:faith}, both assumptions exclude only measure-zero parameter sets, but Assumption \ref{assumption:conv_Faithfulness} may be easier to use in practice. Theorem \ref{thm:DOG} therefore makes the trade-off between assumption strength and identifiability explicit.

Lastly, as shown in Theorem \ref{thm:DOG}, for an LV-SEM, the only undetermined part in the DOG equivalence class pertains to the assignment of the exogenous noises and coefficients, but the structure is the same. Consequently, if only the identification of the structure without weights is of interest, Assumptions \ref{assumption:separability} and \ref{assumption:lv_me_faithfulness} are sufficient.

\begin{remark}
\label{rmk:diff}
We used a different definition for AOG and DOG of a SEM-ME in our preliminary work \citep{yang2022causal2}. In that work, mleaf variables are either assigned to the groups of their (observed or measured) parents or a separate group. In contrast, in this work, mleaf variables cannot be assigned to the groups of their observed parents. This change is based on using the information about which cogent variables are measured and which ones are directly observed. Specifically, 
models in the same equivalence class defined in \citep{yang2022causal2} may have different labeling of $\setY$ and $\setZ$ among variables, 
while in this work, models in the same equivalence class have the same mixing matrix, AOG/DOG and the same labeling of $\setY$ and $\setZ$.
Therefore, this change leads to smaller equivalence classes and hence more identification power.
\end{remark}

\begin{figure}[t!]
\centering
\begin{tikzpicture}[very thick, scale=.32]
\centering
\draw[aog, color=red!20, rounded corners=10] (-5, 8) -- (0.5, -4.5) -- (-8, -5) -- cycle;

\draw[aog, rounded corners=10] (2.5, 7) -- (2.5, -6.5) -- (12.5, 0) -- cycle;

\draw[aog, color=blue!20, rounded corners=10] (-2.5, 6.5) -- (0.5, 4.5) -- (4.5, 10.5) -- (1.5, 12.5) -- cycle;

\foreach \place/\name in  {{(-4,0)/Z_3}}
    \node[observed, label=center:{\color{black} $\name$}] (\name) at \place {};

\foreach \place/\name in  {{(4,0)/Z_7}}
    \node[observed, label=center:{\color{black} $\name$}] (\name) at \place {};

\foreach \place/\name in  {{(0,7)/Z_1}}
    \node[observed, label=center:{\color{black} $\name$}] (\name) at \place {};
    
\foreach \place/\name in  {{(2,10)/Z_2}, {(-6,-3)/Z_4}, {(-4.5,3.5)/Z_5}, {(-2,-3)/Z_6}, {(4.5,3.5)/Z_8}, {(4.5, -3.2)/Z_9}}
    \node[deterministic, label=center:{$ \name$}] (\name) at \place {};

\node[deterministic, label=center:{$Z_{10}$}] (Z_10) at (9, 0) {};

\foreach \source/\dest in {{Z_1/Z_2}, {Z_3/Z_4}, {Z_3/Z_5}, {Z_3/Z_6}, {Z_7/Z_8}, {Z_7/Z_9}}
    \path[causal] (\source) edge (\dest);
    
\foreach \source/\dest in {{Z_1/Z_3}, {Z_1/Z_7}, {Z_3/Z_7}, {Z_7/Z_10}, {Z_3/Z_8}, {Z_3/Z_9}}
    \path[causal] (\source) edge (\dest);
    
\path[causal] (Z_1) edge[in=90, out=0] (Z_10);
\path[causal] (Z_3) edge[in=150, out=15] (Z_10);


\end{tikzpicture}
\hspace{1mm}
\begin{tikzpicture}[very thick, scale=.32]
\centering
\draw[aog, rounded corners=10] (-5, 8) -- (0.5, -4.5) -- (-8, -5) -- cycle;
\fill[myshape, fill=red!20, rounded corners=10] (-5, 8) -- (0.5, -4.5) -- (-8, -5) -- cycle;

\fill[myshape, color=purple!30, rounded corners=10] (2.5, 5.8) -- (2.5, -5.2) -- (6, -5.2) -- (6, 5.8) -- cycle;

\fill[color=orange!30] (9, 0) ellipse (2cm and 2cm);

\draw[aog, line width=0.5pt, rounded corners=10] (2.5, 7) -- (2.5, -6.5) -- (12.5, 0) -- cycle;

\draw[aog, rounded corners=10] (-2.5, 6.5) -- (0.5, 4.5) -- (4.5, 10.5) -- (1.5, 12.5) -- cycle;
\fill[myshape, fill=blue!20, rounded corners=10] (-2.5, 6.5) -- (0.5, 4.5) -- (4.5, 10.5) -- (1.5, 12.5) -- cycle;

\foreach \place/\name in  {{(-4,0)/Z_3}}
    \node[observed, label=center:{\color{black} $\name$}, red!20] (\name) at \place {};

\foreach \place/\name in  {{(4,0)/Z_7}}
    \node[observed, label=center:{\color{black} $\name$}, purple!30] (\name) at \place {};

\foreach \place/\name in  {{(0,7)/Z_1}}
    \node[observed, label=center:{\color{black} $\name$}, blue!20] (\name) at \place {};
    
\foreach \place/\name in  {{(2,10)/Z_2}, {(-6,-3)/Z_4}, {(-4.5,3.5)/Z_5}, {(-2,-3)/Z_6}, {(4.5,3.5)/Z_8}, {(4.5, -3.2)/Z_9}}
    \node[deterministic, label=center:{ $ \name$}] (\name) at \place {};

\node[deterministic, label=center:{$Z_{10}$}] (Z_10) at (9, 0) {};

\foreach \source/\dest in {{Z_1/Z_2}, {Z_3/Z_4}, {Z_3/Z_5}, {Z_3/Z_6}, {Z_7/Z_8}, {Z_7/Z_9}}
    \path[causal] (\source) edge (\dest);
    
\foreach \source/\dest in {{Z_1/Z_3}, {Z_1/Z_7}, {Z_3/Z_7}, {Z_7/Z_10}, {Z_3/Z_8}, {Z_3/Z_9}}
    \path[causal] (\source) edge (\dest);
    
\path[causal] (Z_1) edge[in=90, out=0] (Z_10);
\path[causal] (Z_3) edge[in=150, out=15] (Z_10);

\end{tikzpicture}
\caption{\textbf{Left}: Diagram of the SEM-ME and the corresponding AOG considered in Example \ref{example:ex2}. \textbf{Right}: DOG of the SEM-ME.}
    \label{fig:figure_ex2}
\end{figure}

\begin{example} \label{example:ex2}
Figure \ref{fig:figure_ex2} shows an example of a causal diagram of a SEM-ME with 10 measured variables. $(Z_1, Z_3, Z_7)$ are cogent variables, and the remaining variables are mleaf variables.
The AOG of the model is shown on the left, and the DOG of the model is shown on the right. We note that $Z_{10}$ belongs to the same ancestral ordered group as $Z_7$ since all other parents of $Z_{10}$ are also parents of $Z_7$. However, $Z_{10}$ does not belong to the same direct ordered group as $Z_7$. This is because $Z_8$ is a child of $Z_7$, $Z_1$ is a parent of $Z_{10}$, but $Z_1$ is not a parent of $Z_8$. Therefore the edge $Z_7\to Z_{10}$ violates the Condition \ref{condition:me-id}(b).
\end{example}

\subsection{Identifiability of LV-SEM-ME}
\label{sec:ME-LV-joint}

In this subsection we present identification results for systems in which latent confounding and measurement error coexist. For this general case we require one additional assumption, namely \emph{minimality}, stated below.

\begin{assumption}[Minimality] \label{assumption:minimality}
We assume the linear LV-SEM-ME $M$ is \emph{minimal}, that is, there does not exist any other linear LV-SEM-ME $M'$ such that $M'$ has strictly fewer unobserved variables than $M$, the same observability indicators of the variables, and the same mixing matrix as $M$ up to permutation and scaling of the columns. 
\end{assumption}

The minimality assumption asserts that the ground-truth model has no more unobserved variables than any other model with the same mixing matrix and the same observability indicators. 
This assumption is required since we cannot infer the number of unobserved variables without prior knowledge of the system. Recall from Equation \eqref{eq:Wstar} that the number of columns of $\bW^*$ is the sum of the number of cogent variables and unobserved variables. However, the number of each type of variable is not known a priori under the separability assumption alone.

Minimality assumption is always required when unobserved variables are present in the system. 
Specifically, it is often assumed that the ground-truth model either has the fewest number of edges \citep{adams2021identification}, or the fewest number of unobserved variables \citep{salehkaleybar2020learning}. Our minimality condition falls into the latter case, and
in Proposition \ref{prop:minimality} below, we show that the minimality assumption has an equivalent graphical characterization.

\begin{proposition}[Minimality]
\label{prop:minimality}
Under Assumption \ref{assumption:conv_Faithfulness}, a linear LV-SEM-ME is not minimal if and only if there exists an unobserved variable $H_i$ and a mleaf child $Z_j$ of $H_i$, such that for any other child $V_k$ of $H_i$, $An(Z_j)\subseteq An(V_k)$.\footnote{Note that we do not include $Z$ itself in $An(Z)$.}
This is equivalent to the following. Any (observed or unobserved) parent of $Z_j$ is also an ancestor of $V_k$. 
\end{proposition}

The condition in Proposition \ref{prop:minimality} resembles the condition in Definition \ref{def:ancestral_ordered_group_decomposition} defining AOG. This similarity is not accidental: both characterize situations in which the location of an exogenous source is not identifiable. In Proposition \ref{prop:minimality} the ambiguous source belongs to an unobserved variable; in the AOG definition it belongs to a measured cogent variable.

With the minimality assumption in place, we can state our main identification result for LV-SEM-ME. As in Subsection \ref{sec:ME-LV-sep}, we present separate results under Assumptions \ref{assumption:conv_Faithfulness} and \ref{assumption:lv_me_faithfulness} to highlight the trade-off between assumption strength and identifiability.

\begin{definition}[AOG and DOG of LV-SEM-ME]\label{def:aog_dog_lv_me}
The DOG (resp. AOG) of an LV-SEM-ME consists of a partition of the variables in $\setH\cup\setZ^L\cup\setV^C$ 
described as follows:
\begin{enumerate}[(1)]
\item 
Assign each cogent variable $V_i\in\setV^C$ to a distinct group.

\item
Assign the mleaf variables in $\setZ^L$ to the groups of measured cogent variables or a separate group following (2)(i) in Definition \ref{def:ordered_group_decomposition} (resp. Definition \ref{def:ancestral_ordered_group_decomposition}).

\item 
Assign the unobserved variables in $\setH$ to the groups of cogent variables or a separate group following (2)(ii) in Definition \ref{def:ordered_group_decomposition} (resp. Definition \ref{def:ancestral_ordered_group_decomposition}).
\end{enumerate}
\end{definition}

\paragraph{Graphical characterization}
Using Definition \ref{def:aog_dog_lv_me}, the AOG and DOG equivalence classes of an LV-SEM-ME are defined in the same way as Definitions \ref{def:AOG equivalence class} and \ref{def:DOG equivalence class}, respectively. 
As in Section \ref{sec:ME-LV-sep}, models in the same AOG and DOG equivalence classes are consistent with the different sets of causal orders among the groups, and models in the same DOG equivalence class have the same edges across groups. Proposition \ref{propsition:lvsemme_structure} summarizes the latter fact. 
\begin{proposition}\label{propsition:lvsemme_structure}
Models in the same DOG equivalence class of an LV-SEM-ME have the same unlabeled graph structure, i.e., the causal diagrams of these models are isomorphic.
\end{proposition}
However, since we now consider models with both measured variables and unobserved confounders, unlike the results in Section \ref{sec:ME-LV-sep}, the induced structure on each group may not be a star
graph. Specifically, if there is an edge from the unobserved variables to the mleaf variables in the same group, then the induced structure is not a star graph. Otherwise the structure remains to be a star graph.
We extend our approach by defining the center of a group as the cogent variable in that group (if it exists), or the only mleaf or unobserved variable in the group if the group does not include a cogent variable. In this case, all members in the equivalence classes can be enumerated by either switching the center with any mleaf variables in the group, and/or switching the exogenous noises of the center with the noises of any unobserved variables in the group. For example, a group with one measured cogent variable, one mleaf variable and one unobserved variable has three equivalents (switching the cogent with the mleaf, switching the noise of the cogent with the noise of the unobserved confounder, and both).

We now show that this notion of equivalence is exactly the extent of identifiability in LV-SEM-ME.

\begin{theorem}\label{thm:lv_sem_me}
We have the following results regarding the identification in LV-SEM-ME:
\begin{enumerate}[(a)]
\item Under Assumptions \ref{assumption:separability}, \ref{assumption:conv_Faithfulness} and \ref{assumption:minimality}, the linear LV-SEM-ME can be identified up to its AOG equivalence class.
\item Under Assumptions \ref{assumption:separability}, \ref{assumption:lv_me_faithfulness} and \ref{assumption:minimality}, the linear LV-SEM-ME can be identified up to its DOG equivalence class with probability one.
\end{enumerate}
\end{theorem}

\section{Algorithm}
\label{sec:alg}

In this section, we present recovery algorithms for the introduced LV-SEM-ME model. We first present AOG recovery algorithm (Algorithm \ref{alg:aog-latent}) in Section \ref{sec:aog_recovery}. The algorithm returns the AOG of the underlying model, and it is used in both AOG equivalence class and DOG equivalence class recovery algorithms. 
We then show how to recover all models in the AOG and DOG equivalence classes in Section \ref{sec:dog_recovery} based on the recovered AOG (Algorithm \ref{alg:recovery}).

Both Algorithms \ref{alg:aog-latent} and \ref{alg:recovery} take as input the matrix $\bW^*$ defined in Equation \eqref{eq:Wstar}. 
We note that $\bW^*$ can be recovered from the observational data by first recovering the overall mixing matrix $\bW$ (cf. \eqref{eq:W}) using methods such as overcomplete ICA \citep{eriksson2004identifiability} when the exogenous noises are assumed to be non-Gaussian. Then, $\bW^*$ can be deduced from $\bW$ by removing certain columns as described in Section \ref{sec:assumptions}.

\subsection{AOG Recovery} \label{sec:aog_recovery}
The following property follows directly from the definition of AOG and shows that, under conventional faithfulness, the AOG can be identified from the support pattern of the mixing matrix $\bW^*$ alone.

\begin{proposition} \label{prop:aog_alrorithm}
Under Assumptions \ref{assumption:conv_Faithfulness} and \ref{assumption:minimality}, 
\begin{enumerate}[(a)]
    \item One mleaf variable and one measured cogent variable belong to the same ancestral ordered group if and only if the two rows in $\bW^*$ corresponding to these variables have the same support. 
    Further, for any cogent variable $V_i$ and its descendant $V_j$, the row support of $V_i$ must be a subset of the row support of $V_j$.
    \item  One unobserved variable and one cogent variable belong to the same ancestral ordered group if and only if the two columns in $\bW^*$ corresponding to the exogenous noise terms of these variables have the same support. 
    Further, for any cogent variable $V_i$ and its ancestor $V_j$, the column support of $N_{V_i}$ must be a subset of the column support of $N_{V_j}$.
\end{enumerate}
\end{proposition}

The proof of Proposition \ref{prop:aog_alrorithm} follows directly from the definition of AOG and is therefore omitted.

Equipped with Proposition \ref{prop:aog_alrorithm}, 
we propose an iterative algorithm for recovering the AOG in Definition \ref{def:aog_dog_lv_me} from $\bW^*$. The pseudo-code of the proposed method is presented in Algorithm \ref{alg:aog-latent}.
In the first iteration, the algorithm randomly chooses a row in $\bW^*$ with the fewest non-zero entries and finds all other rows with the same support. Denote the selected rows by $\setZ_\setJ$ and the columns corresponding to these non-zero entries by $\setN_I$. Each selected row may correspond either to a cogent variable or to an mleaf variable, and each column in $\setN_I$ may correspond either to the exogenous noise of a cogent variable or to an unobserved confounder. The task is to decide whether there exists a cogent variable (and its associated exogenous noise). We first select the columns in $\setN_I$ with the fewest non-zero entries in $\bW^*$. Denote this subset by $\setN_J$. Then the noises in $\setN_I\setminus\setN_J$ must correspond to unobserved variables and are assigned to separate groups in $\setC_{unobserved}$.

\begin{algorithm}[t!]
\SetAlgoVlined
\DontPrintSemicolon
\KwIn{Recovered mixing matrix $\bW^*$, observability indicators of the variables (whether each variable is observed or measured).}
Define $\mathbf{n}$ as the vector with the $i$-th entry as the number of non-zero entries in row $i$ of $\bW^*$.\\ Define $\mathbf{m}$ as the vector with the $j$-th entry as the number of non-zero entries in column $j$ of $\bW^*$.
\; 
Initialize $\tilde{\bW}\gets \bW^*$, $\setC_{cogent} \gets [~]$, $\setC_{mleaf} \gets [~]$, $\setC_{unobserved} \gets [~]$. \;
\While{$\tilde{\bW}$ is not empty}{
Find the rows in $\tilde{\bW}$ that contain the fewest number of non-zero entries. Among them, choose one with the lowest  corresponding value in $\mathbf{n}$ (break ties at random).
Denote the selected row by $\bw$, the variable corresponding to this row by $Z_w$, and its corresponding value in $\mathbf{n}$ by $n_0$. \; 
Consider the rows in $\tilde{\bW}$ with the same support (non-zero entries) as $\bw$, including $\bw$ itself. Denote the set of variables corresponding to these rows by $\mathcal{Z}_I$, and denote the noise terms corresponding to the support of $\bw$ by $\setN_I$. \;
Denote the set of variables in $\mathcal{Z}_I$ with corresponding value $n_0$ in $\mathbf{n}$ as $\mathcal{Z}_J$, and the set of noise terms in $\mathcal{N}_I$ with the smallest corresponding value in $\bmm$ as $\mathcal{N}_J$.\;
Assign each variable in $\mathcal{Z}_I\setminus \setZ_J$ to a separate group in $\setC_{mleaf}$. Assign each 
exogenous noise term in $\mathcal{N}_I\setminus \setN_J$ to a separate group in $\setC_{unobserved}$.
\; 
Randomly select one noise term $N_m$ in $\setN_J$. Consider the submatrix $\bW_0$ of $\bW^*$ where the rows correspond to the (column) support of $N_m$, and the columns correspond to the (row) support of $Z_w$.\;
\uIf{~$\bW_0$ includes any zero entry}{Assign each variable in $\setZ_J$ to a separate group in $\setC_{mleaf}$. Assign each 
exogenous noise term in $\setN_J$ to a separate group in $\setC_{unobserved}$.
}
\uElseIf{$\setZ_J$ includes any observed variable}{Assign the observed variable in $\setZ_J$ and all exogenous noise terms in $\setN_J$ to a single group in $\setC_{cogent}$. Assign each remaining variable in $\setZ_J$ to a separate group in $\setC_{mleaf}$.}
\Else{Assign all variables in $\mathcal{Z}_J$ and 
all exogenous noise terms in $\mathcal{N}_J$ to a single group in $\setC_{cogent}$.}
Remove from $\tilde{\bW}$ the rows corresponding to the variables in $\mathcal{Z}_I$, and the columns corresponding to the noise terms in $\setN_I$. Remove the corresponding entries in $\mathbf{m}$ and $\bn$.\;}
\KwOut{$\setC_{unobserved}$, ~$\setC_{cogent}$, ~$\setC_{mleaf}$.}
\caption{AOG Recovery algorithm for linear LV-SEM-ME.}
\label{alg:aog-latent} 
\end{algorithm}

We next check whether any of the rows in $\setZ_\setJ$ can correspond to a cogent variable. If there is an observed variable in $\setZ_\setJ$, then it must be the cogent variable. If all variables are unobserved, then we consider the submatrix $\bW_0$ of $\bW^*$ whose rows correspond to the column support of any variable in $\setN_\setJ$ and whose columns correspond to the row support of $\setZ_\setJ$. If $\setZ_\setJ$ includes a cogent variable, then its corresponding exogenous noise must lie in $\setN_\setJ$, and the remaining noises are unobserved confounders. Since all noises in $\setN_\setJ$ have the same number of non-zero entries, they must have the same support. Moreover, any row that includes this exogenous noise must be a descendant of the cogent variable and, under Assumption \ref{assumption:conv_Faithfulness}, must include all non-zero columns of the cogent variable. This implies that $\bW_0$ contains no zero entry. Therefore, if $\bW_0$ contains a zero entry, then none of the rows in $\setZ_\setJ$ can correspond to a cogent variable. In that case, the rows in $\setZ_\setJ$ and the columns in $\setN_\setJ$ all belong to separate groups in $\setC_{mleaf}$ and $\setC_{unobserved}$, since they correspond to mleaf variables and unobserved variables. If instead $\bW_0$ contains no zero entry, then under the minimality assumption one of the rows in $\setZ_\setJ$ corresponds to a cogent variable. Therefore, all noises in $\setN_J$ and all rows in $\setZ_\setJ$ belong to a single ancestral ordered group in $\setC_{cogent}$.  
The algorithm then removes the rows in $\mathcal{Z}_I$ and the columns in $\setN_I$. Denote the remaining matrix by $\tilde{\bW}$.

In the second iteration, the algorithm again chooses a row $\mathbf{w}$ in $\tilde{\bW}$ with the fewest non-zero entries. However, after the first iteration, a row with the fewest non-zero entries in $\tilde{\bW}$ need not correspond to a cogent variable in the original matrix $\bW^*$, because some non-zero entries may have been removed. Therefore, among the columns in the support of $\mathbf{w}$, we select one column with the fewest non-zero entries in the full matrix $\bW^*$. We then select all other rows in $\tilde{\bW}$ with the same support as $\mathbf{w}$ and denote the resulting set by $\setZ_\setI$. Rows in $\setZ_\setI$ that have more non-zero entries than $\mathbf{w}$ in the full matrix $\bW^*$ must be mleaf variables (otherwise Assumption \ref{assumption:conv_Faithfulness} would be violated) and are assigned to separate groups in $\setC_{mleaf}$. Rows that have the same number of non-zero entries as $\mathbf{w}$ may correspond either to a cogent variable or to an mleaf variable, and we can use the same procedure as in the first iteration to distinguish between them. Repeating this procedure until all variables and noises are assigned yields the full ordered grouping. 

The explanation above implies the identifiability result of Algorithm \ref{alg:aog-latent}, summarized as follows.

\begin{proposition}
    Under Assumptions \ref{assumption:conv_Faithfulness} and \ref{assumption:minimality}, the concatenation of the outputs $(\setC_{unobserved}, \setC_{cogent}, \setC_{mleaf})$ of Algorithm \ref{alg:aog-latent} is the AOG of the LV-SEM-ME corresponding to the input $\bW^*$ in a causal order consistent with the AOG equivalence class of the LV-SEM-ME.
\end{proposition}

\paragraph{Computational complexity} Algorithm \ref{alg:aog-latent} includes $p_c$ steps of iteration, where $p_c$ is the number of cogent variables. Each iteration requires $O(mn)$ calculation and needs $O(mn)$ space, where $m$, $n$ are the dimensions of the mixing matrix $\bW^*$. Recall that $m=p_c + p_{ml}$, and $n=p_c+p_H$, where $p_{ml}$ and $p_H$ stand for the number of mleaf and unobserved variables. Therefore, the total time complexity of the algorithm is $O(p_cmn)$, and the space complexity is $O(mn)$. 

\subsection{Model Recovery in the AOG and DOG Equivalence Class } \label{sec:dog_recovery}
In this section, we present our algorithm for recovering the models in the equivalence class using $\bW^*$. 
The pseudo-code is given in Algorithm \ref{alg:recovery}. 

\begin{algorithm}[t]
\SetAlgoVlined
\LinesNumbered
\DontPrintSemicolon
\KwIn{Recovered mixing matrix $\bW^*$, observability indicators of the variables.}
Obtain the AOG of the true model using Algorithm \ref{alg:aog-latent}. \;

Denote $row$ as a selection of centers in the groups in $\setC_{cogent}$. Define $Row$ as the set of all $row$s.\;
Denote $col$ as a selection of noises in the groups in $\setC_{cogent}$. Define $Col$ as the set of all $col$s.\;
Initialize $\setM_{AOG}=\emptyset$.\;
\For{all $row\in Row$ and $col\in Col$}{
$\bC=\bI - \left(\bW^*[row, col]\right)^{-1}$. \;
$\bB^C=(\bI - \bC) \bW^*[row, col^C]$. \;
$\bD = \bW^*[row^C, col](\bI - \bC)$. \;
$\bB^L= \bW^*[row^C, col^C] - \bD (\bI - \bC)^{-1} \bB^C$.\;
$M = \{\bB=[\bB^L ;\bB^C], \bC, \bD\}$. Add $M$ to $\setM_{AOG}$.\;
}
Define $\setM_{DOG}$ as the set of models in $\setM_{AOG}$ that have the fewest total number of non-zero entries in $\bB$, $\bC$, $\bD$.\;
\KwOut{$\setM_{AOG}$, $\setM_{DOG}$.}
\caption{Recovering all models in the AOG/DOG Equivalence Class} 
\label{alg:recovery} 
\end{algorithm}

\paragraph{AOG Equivalence class}
Recall from Section \ref{sec:ME-LV-joint} that all members in the AOG equivalence class can be enumerated by switching the center with any mleaf variables in the group and/or switching the exogenous noises of the center with the noises of any unobserved variables in the group. Therefore, given the mixing matrix $\bW^*$, Algorithm \ref{alg:recovery} first recovers the AOG of the true model using Algorithm \ref{alg:aog-latent}. Then, it enumerates all possible choices of centers and the noises for each group. Note that groups containing only mleaf variables or noises of unobserved variables (i.e., in $\setC_{mleaf}$ or $\setC_{unobserved}$) only have one choice. Therefore, we only need to consider all the groups that include cogent variables (i.e., in $\setC_{cogent}$). Denote each single selection of the centers in these groups as $row$, and the noises as $col$. The next step is to recover the model parameters $\bB$, $\bC$, $\bD$ based on $\bW^*$ and the selected $row$ and $col$ following Equation \eqref{eq:Wstar}. Denote the variables not in $row$ as $row^C$, and the noises not in $col$ as $col^C$. 
The selected centers in $row$ correspond to $V^C$, and the selected noises in $col$ correspond to $N_{V^C}$ in \eqref{eq:Wstar}. Similarly, variables in $row^C$ and noises in $col^C$ correspond to $Z^L$ and $N_H$, respectively. Therefore $\bC$, $\bB^C$, $\bD$, $\bB^L$ can be calculated following lines 6-9 in Algorithm \ref{alg:recovery}. Finding model parameters $\bB$, $\bC$, $\bD$ for all possible choices of $row$ and $col$ gives us all models in the AOG equivalence class.

\paragraph{DOG Equivalence class}

Proposition \ref{proposition:fewest_edges} allows us to recover models in the DOG equivalence class from the AOG output. It states that the ground-truth model has strictly fewer edges than any model in the same AOG equivalence class that does not belong to the DOG equivalence class.

\begin{proposition}\label{proposition:fewest_edges}
Suppose an LV-SEM-ME satisfies Assumptions \ref{assumption:separability} and \ref{assumption:lv_me_faithfulness}. Any model that belongs to the same AOG equivalence class but does not belong to the same DOG equivalence class has strictly more edges than any member in the DOG equivalence class.
\end{proposition}

Recall from Section \ref{sec:ME-LV-joint} that models in the same DOG equivalence class all have the same unlabeled graph structure, hence the same number of edges. Therefore, using Proposition \ref{proposition:fewest_edges} given the members of the AOG equivalence class of the true model, members in the DOG equivalence class can be found by finding all models in the AOG equivalence class that have the fewest number of edges in the recovery output.

To conclude, the identifiability of Algorithm \ref{alg:recovery} can be summarized by the following proposition.

\begin{proposition}
\begin{enumerate} [(a)]
    \item Under Assumptions \ref{assumption:conv_Faithfulness} and \ref{assumption:minimality},  Algorithm \ref{alg:recovery} outputs $\setM_{AOG}$ which is the AOG equivalence class of the LV-SEM-ME corresponding to the input $\bW^*$.
    \item Under Assumptions \ref{assumption:lv_me_faithfulness} and \ref{assumption:minimality}, Algorithm \ref{alg:recovery} outputs both $\setM_{AOG}$ and $\setM_{DOG}$, where $\setM_{DOG}$ is the DOG equivalence class of the LV-SEM-ME corresponding to the input $\bW^*$.
\end{enumerate}

\end{proposition}

\paragraph{Computational complexity} Recovering the AOG requires $O(p_cmn)$ time and $O(mn)$ space, as discussed above. Recovering a model for a given choice of $row$ and $col$ requires $O(p_cmn)$ time and $O(mn)$ space. Computing the number of edges in a recovered model requires $O(mn)$ time and $O(1)$ additional space.
Denote the total number of choices of the centers and noises (i.e., size of $Row$ and $Col$) by $M_{Row}$ and $M_{Col}$. Note that $M_{Row}$ and $M_{Col}$ are bounded by $|D|^{p_c}$, where $|D|$ is the maximum group size.
Therefore, total time complexity for recovering $\setM_{AOG}$ and $\setM_{DOG}$ is $O(M_{Row}M_{Col}p_cmn)$, and the space complexity is $O(mn)$.

\section{Application to Instrumental Variable, Negative Control, and Front-Door Models}
\label{app:application}

In this section, we show that a small linear LV-SEM, and more generally an LV-SEM-ME, can be used as a common envelope for three classical identification settings: instrumental variables (IV), front-door adjustment, and negative-control outcomes (NCO). Consider the four-node model in Figure~\ref{fig:union_models}(a), where $H$ is unobserved and $Y_1,Y_2,Y_3$ are variables of interest. We write panel~(a) as
\begin{equation*}
    Y_1 = \lambda_1 H + N_1,\qquad
    Y_2 = a_{21}Y_1 + \lambda_2 H + N_2,\qquad
    Y_3 = c_{32}Y_2 + \lambda_3 H + N_3,
\end{equation*}
where $H,N_1,N_2,N_3$ are mutually independent and all displayed coefficients are nonzero unless explicitly restricted. The parameter of interest is the direct causal effect $c_{32}$ from $Y_2$ to $Y_3$. Across the three special cases below, $Y_1$ plays two different classical roles: it is the instrument in the IV submodel and the negative-control outcome in the NCO submodel, while $Y_2$ is the mediator in the front-door submodel.

The three familiar special cases are obtained from panel~(a) by deleting one edge, or by deleting one edge together with a simple coefficient restriction.
\begin{itemize}
    \item \textbf{Instrumental variable model.} Deleting the edge $H\to Y_1$ yields Figure~\ref{fig:union_models}(b), equivalently $\lambda_1=0$. In this case $Y_1$ is independent of the latent confounder $H$, and the only directed path from $Y_1$ to $Y_3$ goes through $Y_2$. Hence
    \begin{equation*}
        c_{32} = \frac{\mathrm{Cov}(Y_1,Y_3)}{\mathrm{Cov}(Y_1,Y_2)}.
    \end{equation*}
    \item \textbf{Front-door model.} Deleting the edge $H\to Y_2$ yields Figure~\ref{fig:union_models}(c), equivalently $\lambda_2=0$. Then $Y_2 = a_{21}Y_1 + N_2$, so the residual of $Y_2$ after regressing on $Y_1$ is $N_2$, which is independent of $(Y_1,H,N_3)$. Therefore the coefficient of $Y_2$ in the population linear regression of $Y_3$ on $(Y_1,Y_2)$ is exactly $c_{32}$.
    \item \textbf{Negative-control-outcome model.} Deleting the edge $Y_1\to Y_2$ and imposing the common-trend restriction $\lambda_1=\lambda_3=\lambda$ yields Figure~\ref{fig:union_models}(d), equivalently $a_{21}=0$ and $\lambda_1=\lambda_3$. Writing the model as
    \begin{equation*}
        Y_1 = \lambda H + N_1,\qquad Y_2 = \alpha H + N_2,\qquad Y_3 = c_{32}Y_2 + \lambda H + N_3,
    \end{equation*}
    we obtain
    \begin{equation*}
        \mathrm{Cov}(Y_2,Y_3) = c_{32}\,\mathrm{Var}(Y_2) + \mathrm{Cov}(Y_1,Y_2),
    \end{equation*}
    and hence
    \begin{equation*}
        c_{32} = \frac{\mathrm{Cov}(Y_2,Y_3)-\mathrm{Cov}(Y_1,Y_2)}{\mathrm{Var}(Y_2)}.
    \end{equation*}
\end{itemize}

\begin{figure}[t!]
\centering
\begin{tabular}{cc}
\begin{tikzpicture}[very thick, scale=0.42]
\foreach \place/\name in {{(-4,0)/Y_1},{(0,0)/Y_2},{(4,0)/Y_3}}
    \node[observed, label=center:{\large $\name$}] (\name) at \place {};
\foreach \place/\name in {{(0,4)/H}}
    \node[latent, label=center:{\large $\name$}] (\name) at \place {};
\foreach \source/\dest in {Y_1/Y_2, Y_2/Y_3, H/Y_1, H/Y_2, H/Y_3}
    \path[causal] (\source) edge (\dest);
\node[] () at (0,-2) {(a) Union model};
\end{tikzpicture}
&
\begin{tikzpicture}[very thick, scale=0.42]
\foreach \place/\name in {{(-4,0)/Y_1},{(0,0)/Y_2},{(4,0)/Y_3}}
    \node[observed, label=center:{\large $\name$}] (\name) at \place {};
\foreach \place/\name in {{(0,4)/H}}
    \node[latent, label=center:{\large $\name$}] (\name) at \place {};
\foreach \source/\dest in {Y_1/Y_2, Y_2/Y_3, H/Y_2, H/Y_3}
    \path[causal] (\source) edge (\dest);
\node[] () at (0,-2) {(b) IV model};
\end{tikzpicture}
\\[2mm]
\begin{tikzpicture}[very thick, scale=0.42]
\foreach \place/\name in {{(-4,0)/Y_1},{(0,0)/Y_2},{(4,0)/Y_3}}
    \node[observed, label=center:{\large $\name$}] (\name) at \place {};
\foreach \place/\name in {{(0,4)/H}}
    \node[latent, label=center:{\large $\name$}] (\name) at \place {};
\foreach \source/\dest in {Y_1/Y_2, Y_2/Y_3, H/Y_1, H/Y_3}
    \path[causal] (\source) edge (\dest);
\node[] () at (0,-2) {(c) Front-door model};
\end{tikzpicture}
&
\begin{tikzpicture}[very thick, scale=0.42]
\foreach \place/\name in {{(-4,0)/Y_1},{(0,0)/Y_2},{(4,0)/Y_3}}
    \node[observed, label=center:{\large $\name$}] (\name) at \place {};
\foreach \place/\name in {{(0,4)/H}}
    \node[latent, label=center:{\large $\name$}] (\name) at \place {};
\foreach \source/\dest in {Y_2/Y_3, H/Y_1, H/Y_2, H/Y_3}
    \path[causal] (\source) edge (\dest);
\node[] () at (0,-2) {(d) NCO model};
\end{tikzpicture}
\end{tabular}
\caption{A four-node union model and its three classical special cases. Panel~(a) is the union model. Panels~(b)--(d) correspond respectively to the restrictions $\lambda_1=0$, $\lambda_2=0$, and $(a_{21}=0,\ \lambda_1=\lambda_3)$.}
\label{fig:union_models}
\end{figure}

The union model in Figure~\ref{fig:union_models}(a) generally violates the graphical assumptions behind all three formulas above. It is not an IV model because the path $Y_1\leftarrow H\to Y_3$ invalidates the instrument; it is not a front-door model because $H\to Y_2$ induces residual confounding between the mediator and the outcome after conditioning on $Y_1$; and it is not an NCO model because $Y_1$ directly affects $Y_2$ and the common-trend restriction need not hold. Consequently, the three classical formulas generally disagree on panel~(a), even though each is correct on its corresponding special case.

When $Y_1,Y_2,Y_3$ are all directly observed, the LV-SEM formulation leads to a singleton DOG equivalence class for each of the four graphs in Figure~\ref{fig:union_models}. In the union model in panel~(a), the three row supports of $\bW^{*}$ are
\begin{equation*}
    \mathrm{Supp}(Y_1)=\{H,N_1\},~
    \mathrm{Supp}(Y_2)=\{H,N_1,N_2\},~
    \mathrm{Supp}(Y_3)=\{H,N_1,N_2,N_3\},
\end{equation*}
so the three observed variables already form distinct ordered groups. The only remaining ambiguity is column-wise: the columns corresponding to $H$ and $N_1$ both have support $\{Y_1,Y_2,Y_3\}$, so the AOG step cannot distinguish them using support alone. Algorithm~\ref{alg:recovery} resolves this ambiguity through the sparsity criterion. Assigning $N_1$ to the first observed group and leaving $H$ as a latent source yields the sparsest compatible model, whereas swapping their roles introduces extra edges. The same argument applies to the front-door submodel in panel~(c), while in panels~(b) and~(d) the relevant columns are already separated by their support patterns.

The fact that the target parameter $c_{32}$ remains identified not only on the IV, front-door, and NCO submodels, but also on the larger union model in panel~(a), can be viewed as a form of \emph{identification robustness}. In the union model, the assumptions behind all three classical formulas may fail simultaneously, yet the parameter of interest is still identified. Thus one does not need to know a priori which of the three specialized designs is correct. Under the broader LV-SEM assumptions used in this paper, the same recovery procedure identifies the causal effect of interest throughout this union family.

An attractive feature of the generalized LV-SEM-ME formulation is that the same template also covers measurement error. Suppose that one of $Y_2$ or $Y_3$ is replaced by a noisy measurement. If $Y_3$ is measured, then the underlying variable becomes an mleaf variable, and its row in $\bW^{*}$ has the same support as the row of $Y_2$. Because $Y_2$ is directly observed, the corresponding direct ordered group has a unique observed center, so the DOG equivalence class remains singleton. If $Y_2$ is measured and $Y_3$ is directly observed, then $Y_2$ is a measured cogent variable and the row supports of the cogent variables remain distinct, so the model is again uniquely identifiable. The only non-singleton case among these simple variants is when both $Y_2$ and $Y_3$ are measured. Then the last direct ordered group contains only measured variables, so the labels within that group are not uniquely determined, although Algorithm~\ref{alg:recovery} still returns the entire DOG equivalence class. Therefore, the example in Figure~\ref{fig:union_models} provides a concrete illustration of how the framework developed in Sections~\ref{sec:id} and~\ref{sec:alg} simultaneously subsumes latent confounding, measurement error, and several classical causal identification designs.

\section{Numerical Experiments}

We report two simulation studies. The first compares our general recovery procedure with estimators tailored to the IV, front-door, and NCO submodels in Figure~\ref{fig:union_models}. The second studies how sensitive the DOG recovery step is to inaccuracies in the mixing matrix.

\subsection{Special-Model Comparison}\label{sec:exp_special_models}

The first experiment is designed to isolate the \emph{second-stage} structural identification problem rather than the \emph{first-stage} estimation of $\bW^{*}$. We used 500 Monte Carlo repetitions in the experiment. For each Monte Carlo repetition and each graph type in Figure~\ref{fig:union_models}, we generated data from a linear model with one latent variable $H$ and three observed variables $(Y_1,Y_2,Y_3)$. The target coefficient $c_{32}$, the edge $Y_1\to Y_2$ when present, and the nonzero loadings of $H$ were drawn independently from $\mathrm{Unif}(0.5,0.9)$. In the NCO model we imposed the restriction $\lambda_1=\lambda_3$, and in the union model we rejected draws with $|\lambda_3-\lambda_1|<0.1$ in order to stay away from the NCO boundary. The exogenous noise terms were generated independently from centered Gamma distributions, which are non-Gaussian and hence compatible with the separability assumption.

\begin{figure}[t!]
    \centering
    \includegraphics[width=0.99\textwidth]{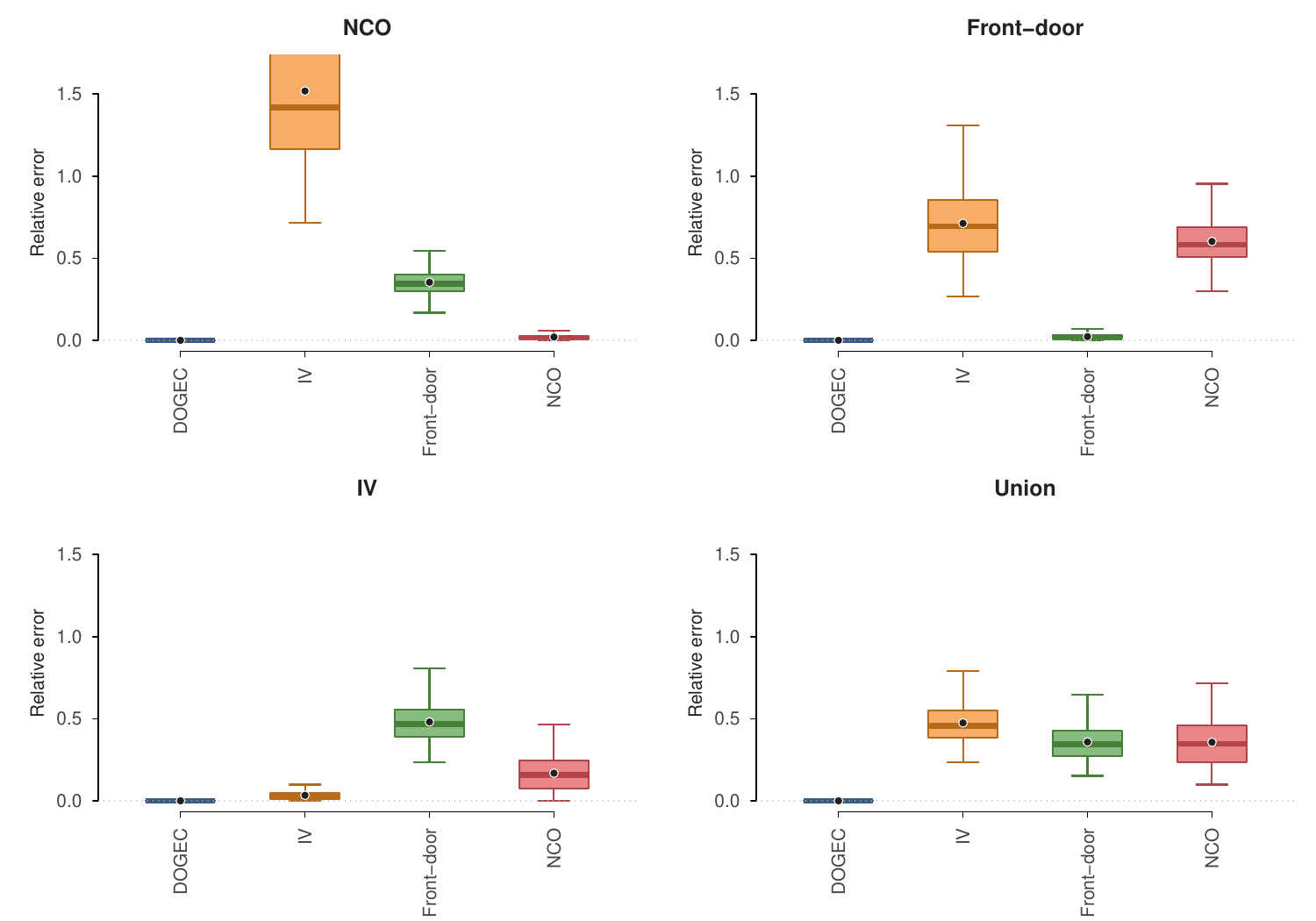}
    \caption{Relative error for estimating $c_{32}$ in the four models of Figure~\ref{fig:union_models}. The DOGEC estimator uses the exact population matrix $\bW^{*}$, so this experiment isolates the second-stage structural recovery problem. The competing methods use the closed-form IV, front-door, and NCO formulas computed from synthetic samples of size $n=4000$.}
    \label{fig:special_models}
\end{figure}

\begin{table}[t!]\small
\centering
\resizebox{\textwidth}{!}{%
\begin{tabular}{l|ccc|ccc|ccc|ccc}
\toprule
& \multicolumn{3}{c|}{\textbf{DOGEC (ours)}} & \multicolumn{3}{c|}{\textbf{IV}} & \multicolumn{3}{c|}{\textbf{Front-door}} & \multicolumn{3}{c}{\textbf{NCO}} \\
& Mean & 20\% & 80\% & Mean & 20\% & 80\% & Mean & 20\% & 80\% & Mean & 20\% & 80\% \\
\midrule
\textbf{NCO} & 0.000 & 0.000 & 0.000 & 1.518 & 1.101 & 1.884 & 0.353 & 0.286 & 0.414 & 0.020 & 0.006 & 0.032 \\
\textbf{Front-door} & 0.000 & 0.000 & 0.000 & 0.712 & 0.513 & 0.891 & 0.022 & 0.007 & 0.036 & 0.602 & 0.490 & 0.713 \\
\textbf{IV} & 0.000 & 0.000 & 0.000 & 0.034 & 0.010 & 0.053 & 0.480 & 0.376 & 0.580 & 0.168 & 0.061 & 0.262 \\
\textbf{Union} & 0.000 & 0.000 & 0.000 & 0.475 & 0.362 & 0.574 & 0.358 & 0.258 & 0.452 & 0.357 & 0.223 & 0.480 \\
\bottomrule
\end{tabular}%
}\caption{Relative error in estimating $c_{32}$ on the four special models.}
\label{tab:special_models}
\end{table}

For the proposed method, we supplied the exact population matrix $\bW^{*}$ to Algorithm~\ref{alg:recovery}. This makes the experiment an oracle second-stage benchmark: the reported performance of our DOG Equivalence Class-based (DOGEC) method reflects only the structural recovery step. We then extracted $c_{32}$ from the recovered DOG representative. As competing methods, we applied the three closed-form estimators motivated in Section~\ref{app:application}: the IV ratio, the front-door linear regression estimator, and the negative-control estimator, each computed from a synthetic sample of size $n=4000$. Therefore, the small nonzero errors of the specialized estimators on their correctly specified submodels are due to ordinary finite-sample estimation, whereas their large errors away from their intended submodels are due to structural misspecification.

Table~\ref{tab:special_models} reports the mean, 20th percentile, and 80th percentile of the relative error $|\hat c_{32}-c_{32}|/|c_{32}|$. Figure~\ref{fig:special_models} shows the same comparison graphically. As expected, each specialized estimator performs well on the model for which it was designed, but its error increases sharply once the corresponding identifying assumptions are violated. In contrast, the proposed LV-SEM-ME recovery procedure remains stable across all four panels because it does not commit to one special graphical pattern a priori. The union model is particularly informative: it is the only setting in which all three specialized estimators are misspecified simultaneously, while the general recovery algorithm still returns the correct effect because the underlying LV-SEM remains identifiable from $\bW^{*}$.

\subsection{Robustness to Noisy and Finite-Sample Mixing Matrices}\label{sec:exp_robustness}

\begin{figure*}[t!]
    \centering
    \includegraphics[width=0.99\textwidth]{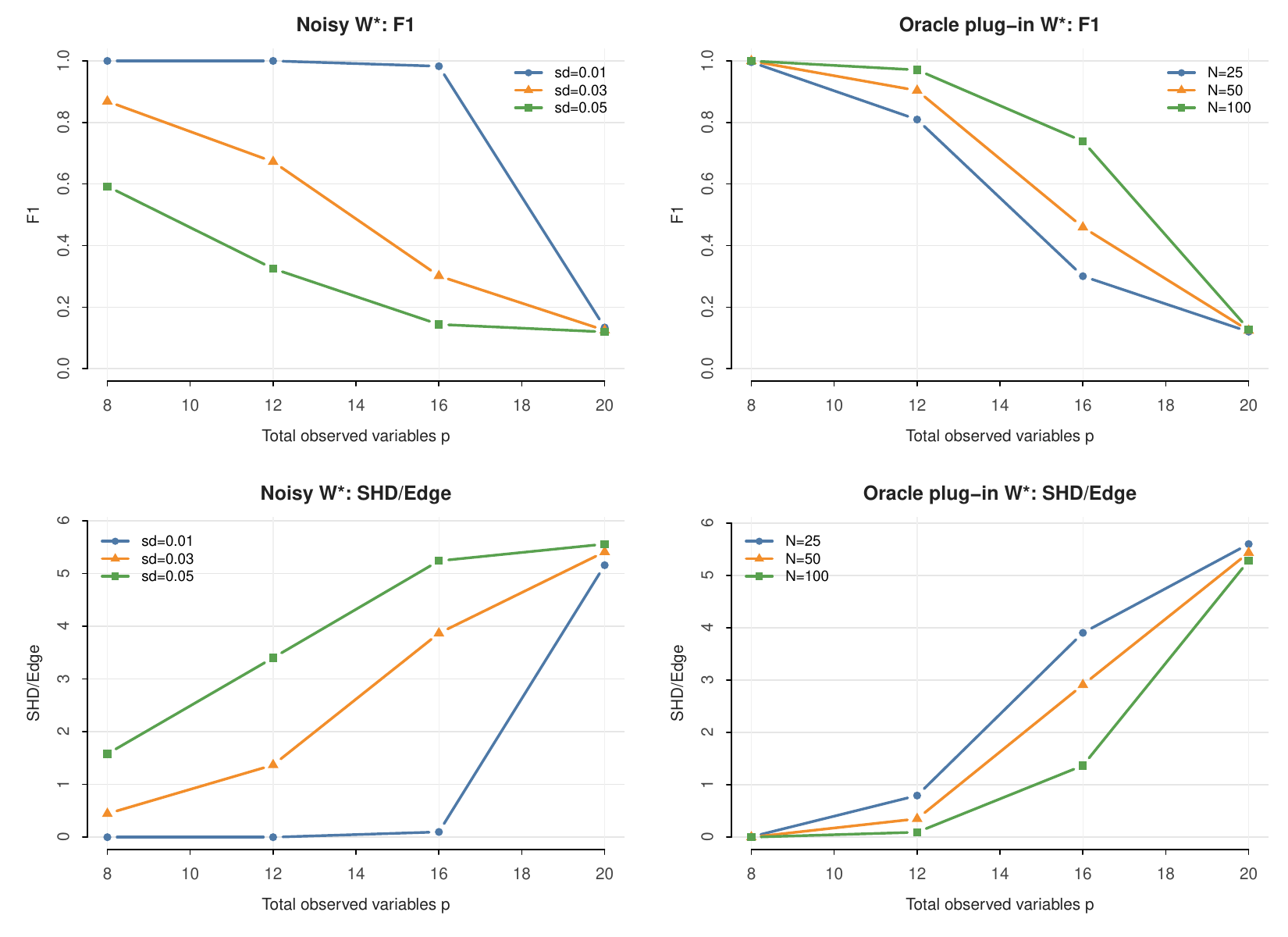}
    \caption{Robustness of DOG recovery on a family of LV-SEM-MEs with both latent confounding and measurement error. Left column: noisy $\bW^{*}$ obtained by perturbing the exact matrix. Right column: oracle plug-in estimate $\hat{\bW}^{*}$ obtained from simulated data by regressing $[X,Y]$ on the simulated sources $[H,N_Y]$, with sample size $n=Np$. The x-axis is the total number of observed variables $p=|X\cup Y|$. Higher F1 and lower SHD/Edge indicate better recovery.}
    \label{fig:robustness}
\end{figure*}

The second experiment studies the sensitivity of the recovery step to inaccuracies in $\bW^{*}$. We considered a deterministic family of canonical LV-SEM-MEs indexed by $q\in\{6,9,12,15\}$. The directly observed cogent variables $Y_1,\ldots,Y_q$ form a directed chain
\begin{equation*}
    Y_1 \to Y_2 \to \cdots \to Y_q,
\end{equation*}
with coefficient $0.8$ on each edge. We set $p_H=p_{ml}=\lfloor q/3\rfloor$. For each $h=1,\ldots,p_H$, a root latent variable $H_h$ affects the two consecutive chain variables $Y_{2h}$ and $Y_{2h+1}$ with coefficient $0.9$. For each $i=1,\ldots,p_{ml}$, we introduced an mleaf variable $Z_i^{L}$ as a child of $Y_{q-p_{ml}+i}$ with coefficient $0.85$, and we observed only its noisy measurement
\begin{equation*}
    X_i = Z_i^{L} + N_{X_i}.
\end{equation*}
Thus the number of directly observed chain variables is $q$, the number of measurements is $p_{ml}$, and the total number of observed variables is $p=q+p_{ml}\in\{8,12,16,20\}$. All exogenous noise terms $H_h$, $N_{Y_j}$, and $N_{X_i}$ were generated independently from centered Gamma distributions. This family was chosen so that the ground-truth DOG equivalence class is a singleton while both latent confounding and measurement error are present.

We used two perturbation mechanisms.
\begin{enumerate}[(i)]
    \item \textbf{Noisy mixing matrix.} Starting from the exact matrix $\bW^{*}$, we first added i.i.d. Gaussian noise $N(0,\delta^2)$ to every nonzero entry of $\bW^{*}$, and then added an independent Gaussian perturbation $N(0,\delta^2)$ to each entry with probability $0.2$, where $\delta\in\{0.01,0.03,0.05\}$. This changes both the numerical values on the true support and, occasionally, the apparent support itself.
    \item \textbf{Oracle plug-in estimate.} For each value of $p$, we generated $n=Np$ observations with $N\in\{25,50,100\}$. Because the experiment is fully synthetic, the simulator knows the source matrix $S=[H,N_Y]$, where $H$ collects the latent root variables and $N_Y$ collects the exogenous noises of the cogent chain variables. Writing the observed data matrix as $O=[X,Y]$, we formed the least-squares estimator
    \begin{equation*}
        \hat{\bW}^{*} = \arg\min_{W}\, \|O-SW^{\top}\|_F^2.
    \end{equation*}
    Equivalently, $\hat{\bW}^{*}$ is obtained by regressing the observed variables on the simulated sources. For the rows associated with the mleaf variables, this is valid because $X_i=Z_i^L+N_{X_i}$ and $N_{X_i}$ is independent of $S$, so the population regression coefficient of $X_i$ on $S$ equals the row of $Z_i^L$ in $\bW^{*}$. This is therefore an \emph{oracle first-stage benchmark}, not a practical estimator, and its purpose is to isolate finite-sample error in the second-stage DOG recovery from approximation error due to a particular ICA implementation.
\end{enumerate}

In both perturbation scenarios, we ran Algorithm~\ref{alg:recovery} on the perturbed matrix using support threshold $0.05$, and we thresholded recovered structural coefficients at $0.30$ when converting the output into an edge set. For each recovered model we then computed the F1 score and the normalized structural Hamming distance (SHD/Edge) between the recovered underlying edge set and the true underlying edge set. Figure~\ref{fig:robustness} reports the averages over 500 Monte Carlo repetitions as a function of the total number of observed variables $p=|X\cup Y|$. The figure shows the expected qualitative pattern: larger perturbations in $\bW^{*}$ and larger graphs make recovery harder, whereas increasing $N$ in the oracle plug-in benchmark improves both metrics. Overall, the results indicate that the second stage of our method is reasonably robust to moderate first-stage error, while also making clear that accurate estimation of the mixing matrix remains as a practical bottleneck.

\section{Conclusion}
\label{sec:conc}
We studied causal discovery in linear systems in which latent confounding and measurement error may coexist. We introduced the LV-SEM-ME model and characterized its identifiability under a separability condition together with two faithfulness assumptions. Under conventional faithfulness, the model is identifiable up to an ancestral ordered grouping (AOG) equivalence class, and under the stronger LV-SEM-ME faithfulness condition, it is identifiable up to the finer direct ordered grouping (DOG) equivalence class. We also gave graphical characterizations of these equivalence classes and recovery algorithms that enumerate all compatible models.

A further contribution of the paper is the identification-robustness perspective developed in Section~\ref{app:application}. The four-node union model shows that instrumental variable, front-door, and negative-control-outcome designs---and simple measurement-error variants of them---can be embedded in a common LV-SEM-ME framework. Within this broader family, the target effect can remain identifiable even when the assumptions underlying the three specialized formulas are not known to hold a priori, and may all fail simultaneously on the union model. In this sense, the framework does not merely recover familiar special cases; it also explains when identification persists beyond them.

The numerical experiments support this perspective from two complementary angles. First, when the population mixing matrix is supplied, the proposed DOG-based recovery procedure accurately recovers the target effect across the instrumental variable, front-door, negative-control, and union settings, whereas the specialized estimators degrade sharply when applied outside their intended submodels. Second, under noisy and finite-sample perturbations of the mixing matrix, the second-stage DOG recovery remains reasonably robust, although performance predictably worsens as the perturbation grows and the graph size increases.

One bottleneck of the proposed methodology remains the first-stage estimation of the mixing matrix. This problem lies outside the scope of the present paper and constitutes an active research area in its own right. Developing more accurate, stable, and practically reliable estimators for the mixing matrix is therefore an important direction for future work. Another natural extension is to relax the linearity assumption and investigate whether analogous identifiability and robustness results can be established in nonlinear settings.

\newpage

\begin{center}
	\textbf{\Large Supplementary Material for ``Causal Discovery in Linear Models with Unobserved Variables and Measurement Error''}

\vspace{5mm}	

Yuqin Yang, Mohamed Nafea, Negar Kiyavash, Kun Zhang,\\ AmirEmad Ghassami
\end{center}

\vspace{15mm}

\appendix
\counterwithin{equation}{section}
\counterwithin{figure}{section}
\counterwithin{table}{section}
\section{Proofs}
\subsection{Proof of Proposition \ref{prop:minimality}}
The proof includes two parts. We first show the sufficiency: For an unobserved variable $H_i$ in the ground truth model $M$, if it has an \uleaf child that satisfies the condition described in Proposition \ref{prop:minimality}, then $M$ is not minimal, i.e., there exists an alternative model $M'$ without $H_i$ that has the same mixing matrix and satisfies Assumption 1.
Next, we show the necessity: If $M$ is not minimal, then there must exist an unobserved variable $H_i$ and one of its mleaf children $Z_j$ such that the described condition is satisfied.

\subsubsection{Proof of sufficiency}

Suppose there exists latent variable $H_i$ and a \uleaf child $Z_j$ of $H_i$ in $M$ such that the condition described in Proposition \ref{prop:minimality} holds. In the following we construct the alternative model $M'$ that includes all variables in $M$ except for $H_i$. The idea is to consider $N_{H_i}$ as the exogenous noise term of $Z_j$. Further, for any other child $V_k$ of $H_i$, replace the edge $H_i\to V_k$ in $M$ by edges from $Z_j$ (and parents of $Z_j$) to $V_k$ in $M'$. 

The structural equation of $Z_j$ in $M$ can be written as
\begin{equation}
\begin{aligned}
    Z_j &= \sum_{l: V_l\in Pa_{M}(Z_j)\cap \setV^C} a_{jl} V_l + \sum_{l_j: H_{l_j}\in Pa_{M}(Z_j)\cap \setH} b_{jl_j} H_{l_j}. 
\end{aligned}
\end{equation}
Consider $N_{H_i}$ as the exogenous noise term of $Z_j$ in $M'$. The structural equation of $Z_j$ in $M'$ is
\begin{equation}
\begin{aligned}
    Z_j &= \sum_{l: Z_l\in Pa_{M}(Z_j)\cap\setV^C} a_{jl} Z_l + \sum_{l_j: H_{l_j}\in Pa_{M}(Z_j)\cap\setH\setminus \{H_i\}} b_{jl_j} H_{l_j} + b_{ji}N_{H_i}. 
\end{aligned}
\label{eq:minimality_2}
\end{equation}

For any other children $V_k$ of $H_i$ in $M$, the structural equation of $V_k$ in $M$ can be written as
\begin{equation}
    V_k = \sum_{l: Z_l\in Pa_{M}(V_k)\cap\setV^C} a_{kl} Z_l + \sum_{l_k: H_{l_k}\in Pa_{M}(V_k)\cap\setH} b_{kl_k} H_{l_k} + N_{V_k}.
    \label{eq:minimality_3}
\end{equation}
By considering $N_{H_i}$ in Equation \eqref{eq:minimality_3} to be the exogenous noise of $Z_j$, the structural equation of $V_k$ in $M'$ can be written as
\begin{equation}
\begin{aligned}
    V_k &= \sum_{l: Z_l\in Pa_{M}(V_k)\cap\setV^C} a_{kl} Z_l + \sum_{l_k: H_{l_k}\in Pa_{M}(V_k)\cap\setH \setminus \{H_i\}} b_{kl_k} H_{l_k} + N_{V_k} + \\
    &~~~~~~~~~b_{ki}b_{ji}^{-1} \left( Z_j - \sum_{l: Z_l\in Pa_{M}(Z_j)\cap\setV^C} a_{jl} Z_l - \sum_{l_j: H_{l_j}\in Pa_{M}(Z_j)\cap\setH\setminus \{H_i\}} b_{jl_j} H_{l_j} \right).
\end{aligned}
\end{equation}

Since $H_i$ and $Z_j$ satisfy the condition in Proposition \ref{prop:minimality}, $V_k$ cannot be an ancestor of $Z_j$ or variables in $Pa_M(Z_j)$ in $M$, otherwise we have $V_k\in \cup_{Z\in Pa(Z_j)}An_M(Z)= An_M(Z_j)\subseteq An_M(V_k)$. This implies that $M'$ is still acyclic. 
Further, since $An_M(Z_j) \subseteq  An_M(V_k)$, there are no additional ancestors introduced to $V_k$ in $M'$ compared with $M$. Lastly, we note that there might be edge cancellations in (4). In particular, the coefficient of the direct edge from a variable in $Z\in Pa_{M}(V_k)\cap\setV^C$ to $V_k$ may change in $M'$ if $Z\in Pa_M(Z_j)$ and hence may be cancelled out. However, $Z$ is still an ancestor of $V_k$ in $M'$, as there is the path $Z\to Z_j\to V_k$. As $M$ and $M'$ has the same mixing matrix, the conventional faithfulness is still satisfied.

In conclusion, if such $H_i$ and $Z_j$ exists in $\setM$, then there exists an alternative model $\setM'$ such that we cannot distinguish $\setM'$ from $\setM$ under Assumption 1 while having one less latent variable. Hence the sufficiency is proved.

\subsubsection{Proof of necessity}
Suppose $M$ is not minimal. Then there exists an alternative model $M'$ that has the same mixing matrix as $M$ and also satisfies conventional faithfulness assumption, while having fewer unobserved variables. Without loss of generality, suppose $M'$ is minimal. Note that since both models correspond to the same mixing matrix, this implies that the number of measured cogent variables in $M$ is strictly less than $M'$, which equals to the number of columns in the mixing matrix minus the number of unobserved variables.

Now, we partition the cogent and mleaf variables in $M$ and $M'$ as follows, where we put measured variables with the same row support in the mixing matrix in the same group. 
We note that in $M'$, this is the same partition as the ancestral ordered grouping among these variables.

Consider the set of measured cogent variables in $M$. According to the definition, they must have different row support hence each of them must belong to a separate group. Therefore, since $M$ has fewer measured cogent variables than $M'$, there exists at least one group $\setG$, where variables $\setG$ are all mleaf variables in $M$ and one of the variables in $\setG$ is a measured cogent variable in $M'$. Denote this measured cogent variable in $M'$ as $Z_j$. Consider the column corresponding to the exogenous noise term of $\setG$ in the mixing matrix.

We first show that this column corresponds to a latent confounder in $M$ by contradiction. Suppose this column corresponds to the exogenous noise of a measured cogent variable $Z_i$ in $M$. Therefore $Z_i$ must be an ancestor of $Z_j$ in $M$, and the entry with row corresponding to $Z_i$ and column corresponding to this noise is not zero.
Denote $Supp(Z_i)$, $Supp(Z_j)$ as the support of the row corresponding to $Z_i$ and $Z_j$, respectively. Since $M$ satisfies conventional faithfulness, the total causal effects from all ancestors of $Z_i$ on $Z_j$ in $M$ is not zero. Hence $Supp(Z_i)\subseteq Supp(Z_j)$. Similarly, consider $Z_i$ in $M'$. Since $Z_j$ is the measured cogent variable, $Z_j$ is an ancestor of $Z_i$, and $Supp(Z_j)\subseteq Supp(Z_i)$. Therefore we have $Supp(Z_j)= Supp(Z_i)$, and hence both belong to the same group $\setG$. However, no variables in $\setG$ are cogent variables in $M$, which leads to a contradiction. Therefore this column must correspond to a latent confounder in $M$; denote that latent confounder by $H_i$.

Next consider any other child $V_k$ of $H_i$ in $M$. Such a variable must be a descendant of $Z_j$, and hence $Supp(Z_j)\subseteq Supp(V_k)$. Because $M$ satisfies conventional faithfulness, this implies $An(Z_j)\subseteq An(V_k)$. Therefore the condition in the proposition holds in $M$. 

\subsection{Enumerating all models in the AOG and DOG equivalence classes by different choice of centers}\label{sec:center}
We first show that an LV-SEM-ME can be uniquely deduced given the mixing matrix $\bW^*$ of the LV-SEM-ME, and a choice of the centers (and their corresponding exogenous noises) in each group. This has been described in Algorithm \ref{alg:recovery}, where the matrices $\bB$, $\bC$, $\bD$ can be found through matrix calculation. 

In this following, given the ground-truth model $M$, we will show how to deduce the structural equations of the variables in the alternative model $M'$, where $M$ and $M'$ have the same mixing matrix and the same (ancestral ordered or directed ordered) grouping of the variables. Specifically, we consider the case when the centers of the groups are the same between $M$ and $M'$ except for one group. We denote the center of this only group in $M$ as $V_i$ with exogenous noise $N_{V_i}$, while the center and the corresponding exogenous noise in $M'$ are $Z_j$ and $N_{H_l}$, where $Z_j$ and $H_k$ belong to the same group as $V_i$ in $M$. We will show the structural equations of all variables that are affected by this difference. This construction is used in the proofs of the AOG and DOG results in Appendices \ref{app:proof_aog} and \ref{app:proof_dog}.

The structural equation of $V_i$ in $M$ can be written as:
\begin{equation}
V_i = \sum_{m_i : H_{m_i} \in Pa_{\setH}(V_i)} b_{i m_i} H_{m_i} + \sum_{n_i : V_{n_i} \in Pa_{\setV^C}(V_{i})} a_{in_i} V_{n_i} + N_{V_i}
\label{eq:V_i_M}
\end{equation}
Note that $H_l\in Pa_{\setH}(V_i)$. For $Z_j$, since it is an mleaf child of $V_i$, we have:
\begin{equation}
Z_j = a_{ji} V_i + \sum_{m_j : H_{m_j} \in Pa_{\setH}(Z_j)} b_{j m_j} H_{m_j} + \sum_{n_j : V_{n_j} \in Pa_{\setV^C}(Z_j)\setminus \{V_i\}} a_{j n_j} V_{n_j}.
\label{eq:Z_j_M}
\end{equation}
We can also write down the equations of any other children $V_{k_i}$ of $V_i$, and any other children $V_{k_l}$ of $H_l$:
\begin{align}
V_{k_i} &= a_{k_i i} V_i + \sum_{m_{k_i} : H_{m_{k_i}} \in Pa_{\setH}(V_{k_i})} b_{k_i m_{k_i}} H_{m_{k_i}} + \sum_{n_{k_i} : V_{n_{k_i}} \in Pa_{\setV^C}(V_{k_i}) \setminus \{V_i\}} a_{k_i n_{k_i}} V_{n_{k_i}} + N_{V_{k_i}},
\label{eq:V_ki_M}\\
V_{k_l} &= b_{k_l l} H_l + \sum_{m_{k_l} : H_{m_{k_l}} \in Pa_{\setH}(V_{k_l}) \setminus \{H_l\}} b_{k_l m_{k_l}} H_{m_{k_l}} + \sum_{n_{k_l} : V_{n_{k_l}} \in Pa_{\setV^C}(V_{k_l})} a_{k_l n_{k_l}} V_{n_{k_l}} + N_{V_{k_l}}.
\label{eq:V_kl_M}
\end{align}

Now, consider $M'$. We can first write down the equation for $V_i$, which is now a mleaf:
\begin{equation}
    V_i = a_{ji}^{-1} \left( Z_j - \sum_{m_j : H_{m_j} \in Pa_{\setH}(Z_j)} b_{j m_j} H_{m_j} - \sum_{n_j : V_{n_j} \in Pa_{\setV^C}(Z_j) \setminus \{V_i\}} a_{j n_j} V_{n_j} \right).
    \label{eq:V_i_new}
\end{equation}
For $Z_j$, by plugging in $V_i$ from \eqref{eq:V_i_M} to \eqref{eq:Z_j_M}, we have:
\begin{equation*}
\begin{aligned}
    Z_j &= a_{ji} \left( b_{i l} H_l +\sum_{m_i : H_{m_i} \in Pa_{\setH}(V_i) \setminus \{H_l\}} b_{i m_i} H_{m_i} + \sum_{n_i : V_{n_i} \in Pa_{\setV^C}(V_i)} a_{i n_i} V_{n_i} + N_{V_i} \right) \\
&~~~+ \sum_{m_j : H_{m_j} \in Pa_{\setH}(Z_j)} b_{j m_j} H_{m_j} + \sum_{n_j : V_{n_j} \in Pa_{\setV^C}(Z_j) \setminus \{V_i\}} a_{j n_j} V_{n_j}.
\end{aligned}
\end{equation*}
Next, since $N_{H_l}$ is the exogenous noise term of $Z_j$, and $N_{V_i}$ is the exogenous noise of $H_l$, we can rewrite it as
\begin{equation}
\begin{aligned}
    Z_j &= a_{ji} \left( H_l +\sum_{m_i : H_{m_i} \in Pa_{\setH}(V_i) \setminus \{H_l\}} b_{i m_i} H_{m_i} + \sum_{n_i : V_{n_i} \in Pa_{\setV^C}(V_i)} a_{i n_i} V_{n_i} \right) \\
&~~~+ \sum_{m_j : H_{m_j} \in Pa_{\setH}(Z_j)} b_{j m_j} H_{m_j} + \sum_{n_j : V_{n_j} \in Pa_{\setV^C}(Z_j) \setminus \{V_i\}} a_{j n_j} V_{n_j} + a_{ji} b_{i l} N_{H_l}.
\end{aligned}
\label{eq:Z_j_new}
\end{equation}
For $V_{k_i}$, by plugging in $V_i$ from \eqref{eq:V_i_new} to \eqref{eq:V_ki_M}, we have
\begin{equation}
\begin{aligned}
V_{k_i} &= \frac{a_{k_i i}}{a_{ji}} \left( Z_j - \sum_{m_j : H_{m_j} \in Pa_{\setH}(Z_j)} b_{j m_j} H_{m_j} - \sum_{n_j : V_{n_j} \in Pa_{\setV^C}(Z_j) \setminus \{V_i\}} a_{j n_j} V_{n_j} \right) \\
&~~~+ \sum_{m_{k_i} : H_{m_{k_i}} \in Pa_{\setH}(V_{k_i})} b_{k_i m_{k_i}} H_{m_{k_i}} + \sum_{n_{k_i} : V_{n_{k_i}} \in Pa_{\setV^C}(V_{k_i}) \setminus \{V_i\}} a_{k_i n_{k_i}} V_{n_{k_i}} + N_{V_{k_i}}.
\end{aligned}
\label{eq:V_ki_new}
\end{equation}
Lastly, for $V_{k_l}$, we substitute $H_l$ in  \eqref{eq:V_kl_M} by $N_{H_l}$ in  \eqref{eq:Z_j_new} and have
\begin{equation}
\begin{aligned}
V_{k_l} &= \frac{b_{k_l l}}{a_{ji} b_{i l}} Z_j - \frac{b_{k_l l}}{b_{i l}} \left( H_l + \sum_{m_i : H_{m_i} \in Pa_{\setH}(V_i) \setminus \{H_l\}} b_{i m_i} H_{m_i} + \sum_{n_i : V_{n_i} \in Pa_{\setV^C}(V_i)} a_{i n_i} V_{n_i} \right) \\
&~~~ - \frac{b_{k_l l}}{a_{ji} b_{i l}} \left( \sum_{m_j : H_{m_j} \in Pa_{\setH}(Z_j)} b_{j m_j} H_{m_j} + \sum_{n_j : V_{n_j} \in Pa_{\setV^C}(Z_j) \setminus \{V_i\}} a_{j n_j} V_{n_j} \right) \\
&~~~+ \sum_{m_{k_l} : H_{m_{k_l}} \in Pa_{\setH}(V_{k_l}) \setminus \{H_l\}} b_{k_l m_{k_l}} H_{m_{k_l}} + \sum_{n_{k_l} : V_{n_{k_l}} \in Pa_{\setV^C}(V_{k_l})} a_{k_l n_{k_l}} V_{n_{k_l}} + N_{V_{k_l}}.
\end{aligned}
\end{equation}
We note that if $V_i\in Pa(V_{k_l})$, then we have to replace $a_{k_li}V_i$ by the right hand side of Equation \eqref{eq:V_i_new}.

To summarize, compared with $M$, the changes in the parent-child relationships among variables in $M'$ can be summarized as follows:
\begin{enumerate}[(i)]
    \item For $V_i$, since it is an mleaf node in $M'$, its parents in $M'$ are the parents of $Z_j$ in $M$ (excluding $V_i$ itself), plus $Z_j$.\label{center_vi}
    \item For $Z_j$, since it is the new center, its parents in $M'$ are the parents of $Z_j$ itself in $M$ (excluding $V_i$), plus the parents of $V_i$ in $M$.\label{center_zj}
    \item For any child $V_{k_i}$ of $V_i$ in $M$ (other than $Z_j$), compared with its parent set in $M$, the new parent set in $M$ replaces $V_i$ by $Z_j$, and includes additional variables that are the parents of $Z_j$ in $M$. \label{center_vki}
    \item For any child $V_{k_l}$ of $H_l$ in $M$ (other than $V_i$), compared with its parent set in $M$, the new parent set in $M'$ additionally includes the new center $Z_j$ and all parents of $Z_j$ in $M'$ (i.e., parents of $V_i$ and $Z_j$ in $M$). \label{center_vkl}
\end{enumerate}
Note that there may be changes of model coefficients if the ``additional variables'' are already in the parent set of $V_{k_i}$ and $V_{k_l}$. In particular, this change may lead to removal of variables from the parent set if the coefficients cancel out each other.

\subsection{Proof of the identification result under conventional faithfulness}\label{app:proof_aog}
In this subsection, we provide the proof of the result regarding the AOG equivalence class in LV-SEM-ME, i.e., Theorem \ref{thm:lv_sem_me}(a). Note that the proof of the result for SEM-ME and LV-SEM, i.e., Theorem \ref{thm:aog}, can be deduced from it. 

To show that the extent of identifiability of an LV-SEM-ME under conventional faithfulness is the AOG equivalence class, we need to show that, for the ground-truth model $M$, any other model $M'$ in the AOG equivalence class of $M$, and any model $M''$ that has the same mixing matrix but does not belong to the AOG equivalence class of $M$:
\begin{enumerate}[label=(1.\alph*)]
    \item $M'$ satisfies conventional faithfulness. \label{aog_1}
    \item $M'$ is consistent with any causal order among the ancestral ordered groups that is consistent with $M$. \label{aog_2}
    \item $M''$ violates conventional faithfulness. \label{aog_3}
\end{enumerate}

Recall that for each cogent variable $V_i$, the ancestral ordered group of $V_i$ includes its mleaf child $Z_j$ if $V_i$ is measured and all other parents of $Z_j$ are also ancestors of $V_i$, and unobserved parent $H_l$ if all other children of $H_l$ are also descendants of $V_i$.

\paragraph{Proof of \ref{aog_1}} 
We note that it suffices to show (a) when the choices of centers (and/or the corresponding exogenous noise) of $M'$ only differs from the choices of $M$ in one group. This is because if there are $p$ differences in the choices of centers, then we can always find a finite sequence of models $M_0\to M_1\to \cdots\to M_p$, where $M_0=M$, $M_p=M'$, and for each $M_{k_p}$, $k_p\in [p]$, the choices of centers only differs from the choices of centers of $M_{k_p-1}$ in one group. If (a) holds for models that differ in only one choice, then by following the sequence of models it also holds for $M_p$. 

We prove by contradiction. Suppose $V$ is an ancestor of $V'$ in $M'$ and the total causal effect from $V$ to $V'$ is zero. Note that the total causal effect from $V$ to $V'$ is equal to the sum of path products from the exogenous noise of $V$ to $V'$. Suppose the exogenous noise of $V$ in $M'$ is the exogenous noise of $V_0$ in $M$. Since $M$ satisfies Assumption \ref{assumption:conv_Faithfulness}, $V_0$ is not an ancestor of $V'$ in $M$. This means that the added edges in $M'$ introduces additional ancestors to $V'$. 

Note that if $V'=Z_j$, then we compare the addition of ancestors to $Z_j$ in $M'$ with the ancestors of $V_j$ in $M$ as both are the center variable in the corresponding model. Similarly, we compare the addition of ancestors to $V_j$ in $M'$ with the ancestors of $V_i$ in $M$.

However, as we described in Appendix \ref{sec:center}, all added edges in $M'$ can be categorized as follows:
\begin{itemize}
    \item If $V'=Z_j$. According to \ref{center_zj}, all added edges in $M'$ are from parents of $Z_j$ in $M$ to $Z_j$. However, parents of $Z_j$ are all ancestors of $V_i$ in $M$. Therefore no additional ancestors are introduced.
    \item If $V'=V_{k_i}$. According to \ref{center_vki}, all added edges in $M'$ are also from parents of $Z_j$ in $M$ to $Z_j$. Since there is $V_i$ is a parent of $V_{k_i}$ in $M$, parents of $Z_j$ are all ancestors of $V_{k_i}$ in $M$.
    \item If $V'=V_{k_l}$. According to \ref{center_vkl}, all added edges in $M'$ are also from parents of $V_i$ or $Z_j$ in $M$ to $Z_j$. Since $V_i$ is an ancestor of $V_{k_l}$ in $M$, parents of $V_i$ or $Z_j$ are all ancestors of $V_{k_l}$.
\end{itemize}
In conclusion, the added edges in $M'$ does not introduce any additional ancestors to $V'$, which leads to a contradiction. Therefore $M'$ satisfies Assumption \ref{assumption:conv_Faithfulness}.

\paragraph{Proof of \ref{aog_2}} Since $M$ and $M'$ both satisfy conventional faithfulness, then Proposition \ref{prop:aog_alrorithm} holds. Further, for any group $g$, define $Supp(g)$ as the row support of variables in $g$ in $\bW^*$ if $g$ includes any observed or measured variables, and if $g=\{H_i\}\subseteq \setH$, define $Supp(g)$ as $\{N_{H_i}\}$. Similarly, define $Supp(N_g)$ as the column support of the exogenous noises of the variables in $g$ in $\bW^*$ if $g$ includes any cogent or unobserved variables, and if $g=\{Z_L\}\subseteq \setZ^L$, define $Supp(g)$ as $\{Z_L\}$. We have the following property.
\begin{proposition}
For two different ancestral ordered groups $g_i$ and $g_j$, the following three conditions are equivalent:
\begin{enumerate}[(i)]
    \item There exists a causal path from one variable in the group of $g_i$ to one variable in the group of $g_j$.
    \item $Supp(g_i)\subset Supp(g_j)$.
    \item $Supp(N_{g_j})\subset Supp(N_{g_i})$.
\end{enumerate}
\end{proposition}

Therefore, given $M$, a causal order among the groups is consistent with the causal order among the variables if and only if the set relations in the mixing matrix hold. Since $M$ and $M'$ have the same mixing matrix and satisfy Assumption \ref{assumption:conv_Faithfulness}, they are consistent with the same set of causal order among the groups.

\paragraph{Proof of \ref{aog_3}}  
Suppose $M''$ does not belong to the AOG equivalence class of $M$. First, consider the number of cogent variables in $M''$, which is equal to the number of columns of $\bW^*$ minus the number of unobserved variables. Given that $M$ is minimal, $M''$ cannot have more cogent variables than $M$. Similarly, if $M''$ has fewer cogent variables than $M$, then $M''$ is not minimal. Therefore, $M$ and $M''$ must have equal number of cogent variables.

Next, consider the cogent variables in $M''$, and their position in the AOG of $M$. Denote these cogent variables as $\setP$. Firstly, note that no two variables in $\setP$ belongs to the same ancestral ordered group of $M$. This is because the mixing matrix corresponding to $\setP$ must be lower-triangular following the causal order, which is impossible when two variables in $\setP$ have the same row support. Therefore, $\setP$ includes at most one variable in each group. Similarly, the exogenous noises of variables in $\setP$ in $M''$, denoted by $N_\setP$, includes the exogenous noise of at most one variable in each group. Suppose $g$ is a group with cogent variables where either (I) $g$ does not include any variable in $\setP$, or (II) $g$ does not include any variable whose corresponding exogenous noise is in $N_\setP$. Note that such a group must exist, otherwise $M''$ belongs to the same AOG equivalence class. Denote this cogent variable in $M$ as $V_i$, and its exogenous noise $N_{V_i}$.

(I): Suppose $g$ does not include any variable in $\setP$. Then $V_i$ is an mleaf in $M''$. Consider $N_{V_i}$ in $M''$. If there exists one parent of $V_i$ in $M''$ that also includes $N_{V_i}$ (i.e., is affected by $N_{V_i}$ directly or indirectly), denote this variable as $V_j$, which must be in $\setP$ and not in $g$. Since $V_j$ includes $N_{V_i}$, it must be a descendant of $V_i$ in $M$. However, since $M$ satisfies conventional faithfulness and $V_j \not\in g$, the row support of $V_j$ must be a superset of $V_i$. This implies that conventional faithfulness is violated on $M''$. 

Therefore, none of the parents of $V_i$ in $M''$ is affected by $N_{V_i}$, and there must be a directed edge from $N_{V_i}$ to $V_i$. Since $V_i$ is an mleaf, $N_{V_i}$ corresponds to a latent confounder in $M''$. Further, for any other children $V$ of $N_{V_i}$ in $M''$, it must be a descendant of $V_i$ in $M$, and hence $Supp(V_i)\subseteq Supp(V)$. According to Proposition \ref{prop:minimality}, this implies that $M''$ is not minimal.

(II): Suppose $g$ does not include any variable whose exogenous noise is in $N_{\setP}$. Then $N_{V_i}$ corresponds to an unobserved confounder in $M''$. Similarly, consider $V_i$ in $M''$. If there is one cogent variable in $M''$ that is a child of $N_{V_i}$ and affects $V_i$ (or $V_i$ is this cogent variable), denote the exogenous noise of this cogent variable as $N_{V}$, which is in $N_\setP$. Since $N_V$ is an ancestor of $V_i$ in $M$, $Supp(N_{V_i})\subseteq Supp(N_{V})$. Further, as $N_V$ is not the exogenous noise of any (observed or latent) variable in $g$, $Supp(N_{V_i})$ must be a strict subset of $Supp(N_{V})$. As any descendants of $N_V$ must also be a descendant of $N_{V_i}$ in $M''$, this implies that conventional faithfulness is violated on $M''$.  

Therefore, none of the parents of $V_i$ in $M''$ is affected by $N_{V_i}$, and $V_i$ cannot be a cogent variable. Hence $V_i$ is an mleaf variable in $M''$ and is directly affected by $N_{V_i}$. Similarly, for any other children $V$ of $N_{V_i}$ in $M''$, it must be a descendant of $V_i$ in $M$, and hence $Supp(V_i)\subseteq Supp(V)$. According to Proposition \ref{prop:minimality}, this implies that $M''$ is not minimal.

\subsection{Proof of the identification result under LV-SEM-ME faithfulness}\label{app:proof_dog}
In this subsection, we provide the proofs of all results regarding the DOG equivalence class in LV-SEM-ME, i.e., Proposition \ref{propsition:lvsemme_structure}, and \ref{proposition:fewest_edges}, and Theorem \ref{thm:lv_sem_me}(b). The corresponding results for SEM-ME and LV-SEM, namely Proposition \ref{propsition:lvsem_structure} and Theorem \ref{thm:DOG}, follow as special cases. 

To show that the extent of identifiability of an LV-SEM-ME under LV-SEM-ME faithfulness is the DOG equivalence class, the proof has two parts. 

First, we show that, for the ground-truth model $M$, any other model $M'$ in the DOG equivalence class:
\begin{enumerate}[label=(2.\alph*)]
    \item $M'$ does not add extra edge compared with $M$.\label{dog_1}
    \item $M'$ does not remove any edge compared with $M$.\label{dog_2}
    \item $M'$ satisfies LV-SEM-ME faithfulness.\label{dog_3}
\end{enumerate}
Therefore we cannot distinguish $M'$ from $M$. Next, any other model $M''$ that is in the AOG equivalence class of $M$ but not the DOG equivalence class:
\begin{enumerate}[label=(2.\alph*), start=4]
    \item $M''$ adds at least one extra edge compared with $M$.\label{dog_4}
    \item $M''$ does not remove any edge compared with $M$, and there is at least one added edge that is not removed.\label{dog_5}
    \item $M''$ violates LV-SEM-ME faithfulness.\label{dog_6}
\end{enumerate}
Therefore we can distinguish $M'$ from $M$.

\subsubsection{Model within the same DOG equivalence class}
Similar to \ref{aog_1} in the AOG proof, we only need to show the result if $M'$ only differs from $M$ in the choices of centers (and/or the corresponding exogenous noise) in one group. If there are $p$ differences in the choices of centers between $M$ and $M'$, then we can always find a finite sequence of models $M_0\to M_1\to \cdots\to M_p$, where $M_0=M$, $M_p=M'$, and for each $M_{k_p}$, $k_p\in [p]$, the choices of centers only differs from the choices of centers of $M_{k_p-1}$ in one group. 

In the following, we show \ref{dog_1} - \ref{dog_3} together for the one difference in the choices of centers. Specifically, for \ref{dog_2}, we show that if one edge is cancelled out, then $M$ violates LV-SEM-ME faithfulness. Further, for \ref{dog_3}, we show that a single change still preserves LV-SEM-ME faithfulness. Following the sequence of the models, all three properties hold for $M'$.

\paragraph{Proof of \ref{dog_1}} 
Following the description in Appendix \ref{sec:center}, after replacing the center $V_i$ in $M$ with $Z_j$, and replacing the exogenous noise $N_{V_i}$ in $M$ with $N_{H_l}$, where $Z_j$ and $H_l$ belong to the same direct ordered group as $V_i$ in $M$, all added edges in $M'$ can be categorized as follows:
\begin{itemize}
    \item For $V_i$: No edges are added.
    \item For $Z_j$: According to \ref{center_zj}, all added edges in $M'$ are from parents of $Z_j$ (excluding $V_i$) in $M$ to $Z_j$. However, according to Condition \ref{condition:me-id}(a), since $Z_j$ belongs to the same direct ordered group as $V_i$, $Pa(Z_j)\setminus \{V_i\}$ is a subset of $Pa(V_i)$. Therefore no edges are added. 
    \item For $V_{k_i}$: According to \ref{center_vki}, all added edges in $M'$ are from parents of $Z_j$ in $M$ (excluding $V_i$) to $V_{k_i}$. Still, according to Condition \ref{condition:me-id}(b), since $Z_j$ belongs to the same direct ordered group as $V_i$, $Pa(Z_j)$ is a subset of $Pa(V_{k_i})$. Therefore no edges are added. 
    \item For $V_{k_l}$: According to \ref{center_vkl}, the added edges are from $Z_j$ and parents of $V_i$ and $V_j$ in $M$ to $V_{k_l}$. Since $H_l$ belongs to the same direct ordered group as $V_i$ in $M$, $\{V_i\}\cap Pa(V_i)\subseteq  Pa(V_{k_l})$ in $M$. Since the center is replaced by $Z_j$ and $Pa(Z_j)\setminus \{V_i\}$ is a subset of $Pa(V_i)$, no edges are added. 
\end{itemize}
Therefore no edges are added to $M'$ compared with $M$.

\paragraph{Proof of \ref{dog_2}} We show that if any edge is cancelled out in $M'$ described in Appendix \ref{sec:center}, then $M$ violates LV-SEM-ME faithfulness. Similarly, all removed edges in $M'$ can be categorized as follows:
\begin{itemize}
    \item For $V_i$: No edges are removed.
    \item For $Z_j$: According to \ref{center_zj}, the edge from $V$ to $V_i$ in $M$ may be cancelled out in $M'$ if $V$ is also a parent of $Z_j$. Suppose the edge from $V$ to $V_i$ is cancelled out in $M'$ because of this. We show that if this happens, then $M$ violates LV-SEM-ME faithfulness. Note that 
    $V$ can be either cogent or unobserved.
    
    \textbf{If $V$ is cogent}: Consider variable $Z_j$, the set of cogent ancestors $J=An_{M'}(Z_j)\cap \setV^C$ in $M'$, and the set of cogent parents $K=Pa_{M'}(Z_j)\cap \setV^C$ in $M'$. We have $An_{M'}(Z_j)\cap \setV^C = An_{V}(V_j)\setminus \{V_i\}$ and $K\subset Pa_V(V_i)$, because $V\not\in K$. Next, following the structural equation of $Z_j$ in $M'$, we have: $Rank(\bW^J_{K\cup \{Z_j\}})=|K|$. However, the minimal bottleneck from $J$ to $K\cup \{Z_j\}$ in $M$ is at least $|K| + 1$, because $K\subset J$, and there is one extra path in $M$ ($V\to V_i \to Z_j$) that is not blocked by $K$. Therefore, $Z_j$ violates Assumption \ref{assumption:lv_me_faithfulness}(b).

    \textbf{If $V$ is unobserved}: Consider the set of cogent ancestors $J=An_{M'}(Z_j)\cap \setV^C$ plus $\{V\}$ in $M'$, and the set of cogent parents $K=Pa_{M'}(Z_j)\cap \setV^C$ in $M'$. In this case, $K$ may equal to $Pa_V(V_i)$. However, we can still show that $Rank(\wzj)=|K|$ (since there is no direct connection from $V$ to $Z_j$ in $M'$), but the minimal bottleneck from $J$ to $K\cup \{Z_j\}$ in $M$ is at least $|K| + 1$, because the path $V\to V_i \to Z_j$ is not blocked. Therefore $Z_j$ still violates Assumption \ref{assumption:lv_me_faithfulness}(b).
    
    \item For $V_{k_i}$: According to \ref{center_vki}, the edge from $V$ to $V_{k_i}$ in $M$ may be cancelled out in $M'$ if $V$ is also a parent of $Z_j$. Suppose the edge from $V$ to $V_{k_i}$ is cancelled out in $M'$ because of this. Similarly, $V$ can be either cogent or observed.

    \textbf{If $V$ is cogent}: Consider variable $V_{k_i}$, $J=An_{V}(\vki)\setminus \{V_i\}\cup \{H_l\}$, and the set of cogent parents $K=Pa_{M'}(V_{k_i})\cap \setV^C$ in $M'$. We have $K\subset PP(\vki)$ (note that $Z_j$ is an mleaf in $M$), and $|K|< Pa_V(V_{k_i})$. Further, $J$ includes all variables with the exogenous noise corresponding to the cogent ancestors of $Z_j$ in $M'$.
    Therefore, we have $Rank(\wki) =|K|$. However, the minimal bottleneck from $J$ to $K\cup \{\vki\}$ in $M$ is at least $|K| + 1$. Firstly, $K\setminus \{Z_j\}$ is a subset of $J$ so they are included in any bottleneck. Additionally, there are two distinct paths from $J$ to $K\cup \{\vki\}$ that cannot be blocked by $K\setminus \{Z_j\}$: $V\to \vki$ and $H_l\to V_i\to Z_j$. Therefore $\vki$ violates Assumption \ref{assumption:lv_me_faithfulness}(a).

    \textbf{If $V$ is unobserved}: Similarly, consider $J=An_{V}(\vki)\setminus \{V_i\}\cup \{V, H_l\}$ and $K=Pa_{M'}(V_{k_i})\cap \setV^C$. Note that $V\neq H_l$ as $V$ is a parent of $\vki$ in $M$.  
    The results above still hold, as $Rank(\wki) =|K|$, and the same two paths, $V\to \vki$ and $H_l\to V_i\to Z_j$, cannot be blocked by $K\setminus \{Z_j\}$. Therefore $\vki$ violates Assumption \ref{assumption:lv_me_faithfulness}(a).
    
    \item For $\vkl$: According to \ref{center_vkl}, the edge from $V$ to $\vkl$ in $M$ may be cancelled out in $M'$ if $V=Z_j$, or $V\in Pa_{M'}(Z_j)\cap V^C$. Suppose the edge from $V$ to $\vkl$ is cancelled out in $M'$ because of this.

    \textbf{If $V$ is cogent and $V\in Pa_V(V_i)$}: Consider variable $V_{k_l}$, the set of cogent ancestors $J=An_{V}(\vkl)\setminus \{V_i\}\cup \{H_l\}$ in $M$, and the set of cogent parents $K=Pa_{M'}(\vkl)\cap \setV^C$ in $M'$. Similarly, we have $K\subset PP(\vkl)$, and $|K|< Pa_V(V_{k_l})$. We have $Rank(\wkl) =|K|$. Further, the minimal bottleneck from $J$ to $K\cup \{\vkl\}$ in $M$ is at least $|K| + 1$. Firstly, $K\setminus \{Z_j\}$ is a subset of $J$ so they are included in any bottleneck. Additionally, there are two distinct paths from $J$ to $K\cup \{\vkl\}$ that cannot be blocked by $K\setminus \{Z_j\}$: $V\to \vkl$ and $H_l\to V_i\to Z_j$. Therefore $\vkl$ violates Assumption \ref{assumption:lv_me_faithfulness}(a).

    \textbf{If $V=Z_j$}: Consider the same $J$ and $K$ as above. The main difference here is that $Z_j\not\in K$ and hence $K\subset J$. Therefore, $Rank(\wkl) =|K|$, and the minimal bottleneck at least includes all variables in $K$, and one extra variable on the edge $H_l\to \vkl$. Therefore $\vkl$ violates Assumption \ref{assumption:lv_me_faithfulness}(a).

    \textbf{If $V$ is unobserved}: Consider $J=An_{V}(\vkl)\setminus \{V_i\}\cup \{H_l,V\}$ in $M$, and $K=Pa_{M'}(\vkl)\cap \setV^C$ in $M'$. We have $Rank(\wkl) =|K|$, and the minimal bottleneck includes all variables in $K\setminus \{Z_j\}$, as well as two variables on the paths $V\to \vkl$ and $H_l\to V_i\to Z_j$. Therefore $\vkl$ violates Assumption \ref{assumption:lv_me_faithfulness}(a).
\end{itemize}

In conclusion, if an edge is cancelled in $M'$, then $M$ violates Assumption \ref{assumption:lv_me_faithfulness}.

\paragraph{Proof of \ref{dog_3}} From \ref{dog_1} and \ref{dog_2}, we show that $M'$ has the same unlabeled graph structure as $M$. We can construct a mapping $\sigma_M$ on variables in $\setZ\cup\setY$, where 
\begin{equation*}
    \sigma(V_i) = Z_j;\quad \sigma(Z_j) = V_i;\quad \sigma(V) = V, \quad V \neq V_i, Z_j.
\end{equation*}
Similarly, define a mapping $\sigma_N$ on variables in $\setV^C\cup \setH$, where
\begin{equation*}
    \sigma_N(V_i) = H_l;\quad \sigma(H_l) = V_i;\quad \sigma(V) = V, \quad V \neq V_i, H_l.
\end{equation*}

We note that for all variable $V$ except $V_i$ and $Z_j$, $PP_M(V) = PP_{M'}(V)$, $An_{M'}(V)=An_M(V)$. This is because $PP(V)$ either includes both $V_i$ and $Z_j$, or include neither of them, as all parents of $Z_j$ in $M$ are ancestors of $V_i$. Similarly, $An_M(V)$ either includes both $H_l$ and $V_i$, or neither of them, because $H_l$ is a parent of $V_i$, and all other children of $H_l$ are descendants of $V_i$.

In the following, we show that if $M$ satisfies Assumption \ref{assumption:lv_me_faithfulness}, then $M'$ also satisfies Assumption \ref{assumption:lv_me_faithfulness} with probability one. In particular, following the rewriting procedure described in Appendix \ref{sec:center}, we can construct an invertible mapping between the model parameters in $M$ and in $M'$. Since Assumption \ref{assumption:lv_me_faithfulness} is violated with probability zero on the model parameters in $M$, the same results hold on the model parameters in $M'$.

We note that the differences in the model parameters that are different between $M$ and $M'$ can be summarized as follows:
\begin{enumerate}[(i)]
    \item The edge weight from $Z_j$ to $V_i$ is the inverse of the edge weight from $V_i$ to $Z_j$, $a_{ji}$;
    \item The edge weight from other parents of $V_i$ to $V_i$ in $M'$ can be written as the edge weight from other parents of $Z_j$ to $Z_j$ in $M$ multiplied by $- a_{ji}^{-1}$;
    
    \item The edge weight from $Pa(Z_j)$ to $Z_j$ in $M'$ can be written as a function of $a_{ji}$, the edge weight from $Pa(V_i)$ to $V_i$ in $M$ and the edge weight from $Pa(Z_j)$ to $Z_j$;

    \item The edge weight from $Z_j$ to $\vki$ is edge weight from $V_i$ to $\vki$, $a_{k_ii}$ divided by $a_{ji}$;
    \item The edge weight from other parents of $\vki$ to $\vki$ in $M'$ can be written as a function of $a_{k_ii}$, $a_{ji}$ the edge weight from $Pa(\vki)$ to $\vki$ and the edge weight from $Pa(Z_j)$ to $Z_j$ in $M$;

    \item The edge weight from $H_l$ to $\vkl$ is edge weight from $H_l$ to $\vkl$, $b_{k_ll}$, divided by $b_{il}$ (note that $b_{il}$ is included in (iii));
    \item The edge weight from $Z_j$ to $\vkl$ is a function of the edge weight from $V_i$ to $\vkl$ ($a_{k_li}$), $a_{ji}$, $b_{il}$, $b_{k_ll}$;
    \item The edge weight from $Pa(\vkl)$ to $\vkl$ in $M'$ can be written as a function of the edge weight from $Pa(\vkl)$ to $\vkl$, and the edge weight from $Pa(Z_j)$ to $Z_j$, and $Pa(V_i)$ to $V_i$ in $M$.
\end{enumerate}

Therefore, if we arrange the model parameters in $M$ and in $M'$ following (i)-(viii) above, we can clearly see that the mapping that translates model parameters in $M$ to model parameters in $M'$ is invertible. This is consistent with the fact that since $M$ and $M'$ belong to the same DOG, we can equivalently write the model parameters in $M$ using model parameters in $M'$. Therefore, since the model parameters in $M$ satisfy Assumption \ref{assumption:lv_me_faithfulness} with probability one, the model parameters in $M'$ also satisfy Assumption \ref{assumption:lv_me_faithfulness} with probability one.

Lastly, we note that, since the number of models in the DOG equivalence class is finite, if the ground-truth model satisfies Assumption \ref{assumption:lv_me_faithfulness}, then the probability that models in the DOG equivalence class all satisfy Assumption \ref{assumption:lv_me_faithfulness} is also one.

\subsubsection{Model outside the DOG equivalence class}
Without loss of generality, we only need to prove the result in the case where all differences in the choices of centers between $M''$ and $M$ lie outside the corresponding direct ordered group of the cogent variable in $M$, but remain within the same ancestral ordered group. That is, we take $M$ to be the model that is closest to $M''$ in terms of the center choices. We have shown above that this closest model also satisfies LV-SEM-ME faithfulness and has the same unlabeled graph structure.

\paragraph{Proof of \ref{dog_4}} We first show that if $M''$ differs from $M$ in only one choice of center, then at least one edge is added in $M''$. The proof follows the procedure described in Appendix \ref{sec:center}. If $Z_j$ belongs to the same ancestral ordered group as $V_i$ but not to the same direct ordered group, then Condition \ref{condition:me-id} is satisfied. In that case, there is one additional edge either from a parent of $Z_j$ in $M$ to $Z_j$, or to $V_{k_i}$. Similarly, if $H_l$ belongs to the same ancestral ordered group as $V_i$ but not to the same direct ordered group, then Condition \ref{condition:ur_id_new} is satisfied. In that case, there is one additional edge from $Z_j$, or from a parent of $Z_j$ or $V_i$, to $\vkl$. Therefore, $M''$ has at least one more edge than $M$.

If there are multiple differences in the choice of centers between $M''$ and $M$, then the above one-step analysis does not apply directly. In particular, because one added edge may change parent-child relations among variables, it may happen that no edge is added when passing from $M_1$ to $M_2$. Nevertheless, we can still show that $M_2$ has at least one more edge than $M_0$. Repeating this argument along the chain $M_0\to M_1\to \cdots\to M_p$ completes the proof. 

\paragraph{Proof of \ref{dog_5}} We note that we can use the same method as in \ref{dog_2} to show that if any edge is removed in $M$, then $M$ violates Assumption \ref{assumption:lv_me_faithfulness}. Specifically, in \ref{dog_2}, we only used the fact that $Z_j$ is a child of $V_i$, and $H_l$ is a parent of $V_i$. Both still hold true for AOG.
In the following, we will only show the latter part, that is, there is at least one added edge that is not removed.

We prove by contradiction. Suppose all added edges are removed in $M''$, and suppose the edge from $V_0$ to $V$ is added but finally removed. We further assume that among all the added edges $V_1\to V_2$, $V_0$ has the smallest index (following the causal order in $M$), and for any $V'\in De(V_0)\cap An(V)$, no edges from $V_0$ to $V'$ is added. That is, for any causal path from $V_0$ to $V$, no edges are added from $V_0$ to any other variable $V'$ on this path.
Note that $V_0$ and $V$ refers to their position (center or non-center) in $M$, i.e., they may represent different variables in $M''$ and $M_{q}$.

Suppose $M_{s}$ is the first model following the sequence where this edge is added, and $M_{q-1}$ is the last model following the sequence where this edge is present. Denote the center of the ancestral ordered group 
that changes between $M_{s-1}$ and $M_s$ as $V_{i_s}$, and between $M_{q-1}$ and $M_q$ as $V_{i_q}$. Since the edge from $V_0$ to $V$ is added in $M_s$, it must belong to one of the following three cases: There is a causal path $V_0\to V_{i_s}\to V$, or there exists a latent confounder with $V_0 \to V_{i_s} \gets H_{l_s} \to V$ (note that $V_0$ may be $V_{i_s}$) or $V_0 \to Z_{j_s} \gets V_{i_s} \gets H_{l_s} \to V$ in $M_{q-1}$. Note that in the latter two cases, since $Z_{j_s}$, $V_{i_s}$, $H_{l_s}$ all belong to the same ancestral ordered group, $V_0$ is an ancestor of $V_{i_s}$, and $V_{i_s}$ is an ancestor of $V$.

We consider the following cases:

\begin{itemize}
    \item $V=Z_{j_q}$. This step does not include edge removals.
    \item $V=V_{i_q}$. Since the edge is removed in $M_q$, $V_0\in Pa_{M_{q-1}}(Z_{j_q})$ in $M_{q-1}$, where $V_0$ is not a parent of $V_{i_q}$ in $M$, and is not a parent of $Z_{j_q}$ in $M''$. We note that the edge from $V_0$ to $Z_{j_q}$ cannot be an added edge. Because this edge is not removed in $M_q$, and will never be removed for all $M_{q+1}$, $\cdots$, $M_p=M''$. Therefore this edge is in $M$.

    Consider $Z_{j_q}$, and consider $J$ as the set of (unobserved or cogent) variable $V'$ in $M$ where $N_{V'}$ corresponds to the exogenous noise of a variable in $\{V_0\}\cup (An_{M''}(Z_{j_q})\cap \setV^C) $. In other words, $J= An_{M''}(Z_{j_q})\cap \setV^C $ if $V_0$ is a cogent variable, and $J$ additionally includes $V_0$ if it is unobserved. Further, consider $K = Pa_{M''}(Z_{j_q})\cap \setV^C$. It follows that $K\subseteq PP_M(V_{i_q})$, and $J\subseteq An_{M}(Z_{j_q})\setminus \{V_{i_q}\}$. 

    We have $Rank(\bW^J_{K\cup \{Z_{j_q}\}}) = |K|$. However, note that we can find $K$ distinct paths from variables in $J$ to variables in $K$. Specifically, for each variable $V'$ in $K$, define $f(V')$ as the variable in $J$ whose exogenous noise in $M$ is the exogenous noise of $V'$ in $M''$. Note that $f(V')=V'$ if there is no change on $V'$ and $N_{V'}$ on both models. Further, $f(V')$ and $V'$ must belong to the same ancestral ordered group and hence there is a causal path $f(V') \leadsto V'$ in $M$. 
    Therefore, a minimal bottleneck from $J$ to $K\cup \{Z_{j_q}\}$ must at least include $K$ variables. However, there is still one path $V_0\to Z_{j_q}$ that is not blocked, as $V_0$ is not a parent of $Z_{j_q}$ in $M''$ (meaning $V_0\not\in K$). Hence $M$ violates Assumption \ref{assumption:lv_me_faithfulness}.

    \item $V=\vki$ for some $\vki$. Note that this variable $\vki$ may have different positions (center or non-center) in $M_q$ and $M''$. For simplicity of notation, denote this variable as $V_{k_q}$. Then there exist paths $V_0\to V_{i_q}\to V_{k_q}$, $V_0\to V_{k_q}$, and $V_0\to Z_{j_q}$ in $M_{q-1}$. Note that the edge $V_{i_q}\to V_{k_q}$ and $V_0\to Z_{j_q}$ are in $M$.
    
    \textbf{If $V_{k_q}$ does not change the position.} 
    Consider $J$ as the set of (unobserved or cogent) variable $V'$ in $M$ where $N_{V'}$ corresponds to the exogenous noise of a variable in $\{V_0\}\cup (An_{M''}(V_{k_q})\cap \setV^C )$, and $K = Pa_{M''}(V_{k_q})\cap \setV^C$. It follows that $K\subseteq PP_M(V_{k_q})$, and $J\subseteq An_{M}(V_{k_q})$.
    Further, $Rank(\bW^J_{K\cup \{V_{k_q}\}}) = |K|$. However, a minimal bottleneck from $J$ to $K\cup \{V_{k_q}\}$ must at least include: One variable between $V'$ and $f(V')$ for each $V'\in K$, except for $Z_{j_q}$; One variable on the edge $V_{i_q} \to V_{k_q}$; One variable on the edge $V_0\to Z_{j_q}$. Therefore the size of the minimal bottleneck is at least $|K|+1$. Hence $M$ violates Assumption \ref{assumption:lv_me_faithfulness}.

    \textbf{If $V_{k_q}$ changes the position from non-center to center.} That is, it can be denoted by $Z_{j_r}$ for some $ q<r\leq p$. Note that the reason we consider $M''$ in the above analysis is because there may still be edge additions or removals on $V_{k_q}$ after $M_q$. However, if $V=Z_{j_r}$, then we know that there will be no edge additions/removals on $V$ after $M_r$. Therefore, consider $J$ as the set of variable $V'$ in $M_r$ where $N_{V'}$ corresponds to the exogenous noise of a variable in $\{V_0\}\cup (An_{M_{r}}(Z_{j_r})\cap \setV^C )$, and $K = Pa_{M_{r}}(Z_{j_r})\cap \setV^C$. This implies that $J\subseteq An(Z_{j_r})\setminus \{V_{i_r}\}$, and $K\subseteq PP(V_{i_r})$. We have $\ww{Z_{j_r}}= |K|$. However, a minimal bottleneck from $J$ to $K\cup \{V_{k_q}\}$ must at least include: One variable between $V'$ and $f(V')$ for each $V'\in K$, except for $Z_{j_q}$; One variable on the edge $V_{i_q} \to V_{k_q}$; One variable on the edge $V_0\to Z_{j_q}$. Therefore the size of the minimal bottleneck is at least $|K|+1$. Hence $M$ violates Assumption \ref{assumption:lv_me_faithfulness}.

    \textbf{If $V_{k_q}$ changes the position from center to non-center.} That is, it can be denoted by $V_{i_r}$ for some $ q<r\leq p$. Similarly, there are no edge additions or removals involving $V_{i_r}$ after $M_r$. In this case, we consider the variable $V_{i_r}$ in $M_{r-1}$, i.e., $J$ as the set of variable $V'$ in $M_r$ where $N_{V'}$ corresponds to the exogenous noise of a variable in $\{V_0\}\cup (An_{M_{r-1}}(V_{i_r})\cap \setV^C )$, and $K = Pa_{M_{r-1}}(V_{i_r})\cap \setV^C$. The following is the same as when $V_{k_q}$ does not change the position, and we can conclude that $M$ violates Assumption \ref{assumption:lv_me_faithfulness}.

    \item $V=\vkl$ for some $\vkl$. Similarly, for notation simplicity, denote this variable as $V_{l_q}$. Note that this is different from $H_{l_q}$, which refers to the unobserved variable that is in the same group as $V_{i_q}$. Then there exist paths $V_0 \to V_{i_q} \gets H_{l_q} \to V_{l_q}$ or $V_0 \to Z_{j_q} \gets V_{i_q} \gets H_{l_q} \to V_{l_q}$ in $M_{q-1}$.
    
    \textbf{If $V_{l_q}$ does not change the position.} Consider $J$ as the set of variable $V'$ in $M$ where $N_{V'}$ corresponds to the exogenous noise of a variable in $\{V_0\}\cup (An_{M''}(V_{l_q})\cap \setV^C )$, and $K = Pa_{M''}(V_{l_q})\cap \setV^C$. It follows that $K\subseteq PP_M(V_{l_q})$, and $J\subseteq An_{M}(V_{l_q})$, and $\ww{V_{l_q}}=|K|$.

    \begin{itemize}
        \item If $V_{i_q}$ ($Z_{j_q}$ if the center is replaced) is not a parent of $V_{l_q}$ in $M''$. Note that this includes the case when $V_0=V_{i_q}$. Similar to the above analysis for $\vki$, for each variable $V'$ in $K$, define $f(V')$ as the variable in $J$ whose exogenous noise in $M$ is the exogenous noise of $V'$ in $M''$. Then any bottleneck from $J$ to $K$ must include at least one variable between $V'$ and $f(V')$ for each $V'\in K$. Further, it must also include one variable on the edge $H_{l_q}\to V_{l_q}$. Therefore the size of the minimal bottleneck must be at least $|K|+1$.
        \item If $V_{i_q}$ ($Z_{j_q}$ if the center is replaced) is a parent of $V_{l_q}$ in $M$. Then any bottleneck from $J$ to $K$ must include at least one variable between $V'$ and $f(V')$ for each $V'\in K$ except for $V_{i_q}$. Further, it must also include one variable on the edge $H_{l_q}\to V_{l_q}$, and one variable on the edge $V_0\to V_{i_q}$ (or $V_0\to Z_{j_q}$ if $Z_j$ is the center node in $M''$). Therefore the size of the minimal bottleneck must also be at least $|K|+1$.
    \end{itemize}

    Therefore, in both cases, $M$ violates Assumption \ref{assumption:lv_me_faithfulness}.

    \textbf{If $V_{l_q}$ changes its position.} Similar to the analysis described in $\vki$, $V_{l_q}$ can be denoted by $Z_{j_r}$ or $V_{i_r}$ for some $ q<r\leq p$. Since there are no edge additions or removals involving $V_{i_r}$ after $M_r$, we can consider the sets $J$ and $K$ defined on $M_{r-1}$ if $V=V_{i_r}$, and the sets $J$ and $K$ defined on $M_{r}$ if $V=Z_{j_r}$. Following the same analysis as above, we can conclude that $M$ violates Assumption \ref{assumption:lv_me_faithfulness}.
    
\end{itemize}

In conclusion, we show that if all added edges in $M_1,\cdots, M_p$ are removed, then $M$ violates Assumption \ref{assumption:lv_me_faithfulness}. Therefore at least one of the added edge to $M''$ is not removed.

\paragraph{Proof of \ref{dog_6}} Lastly, we show that because of this added edge, $M''$ violates Assumption \ref{assumption:lv_me_faithfulness}. Suppose the edge from $V_0$ to $V$ is added in $M''$. Note that this edge does not introduce any additional ancestral relations as $M''$ and $M$ belong to the same AOG equivalence class. Therefore for each center or non-center variable $V$, the ancestor set and possible parent set of $V$ remain the same. The proof of this claim is actually the same as \ref{dog_2}, but in a reversed manner. 
\begin{itemize}
    \item $V=Z_j$. Then $V$ is a center node in $M''$, and $V_0\in Pa_M(Z_j)\setminus Pa_M(V_i)$.
    Consider $V_i$ in $M''$, which is an mleaf. Consider $J$ as the set of (unobserved or cogent) variable $V'$ in $M''$ where $N_{V'}$ corresponds to the exogenous noise of a variable in $\{V_0\}\cup (An_{M}(V_{i})\cap \setV^C)$, and consider $K = Pa_{M}(X_{i})\cap \setV^C$. We have $J\subseteq An_{M''}(V_i\setminus \{Z_j\})$, and $K\subseteq PP_{M''}(V_j)$. Further, according to the structural equation of $V_i$ in $M$, $\ww{V_i}=|K|$. 
    
    For each variable $V'$ in $K$, define $f(V')$ as the variable in $J$ whose exogenous noise in $M''$ is the exogenous noise of $V'$ in $M$. Then the minimal bottleneck must include at least one variable between $V'$ and $f(V')$ for each $V'\in K$, and one variable on the path $V_0\to Z_j\to V_i$ on $M''$. Therefore the size of the minimal bottleneck must be at least $|K|+1$.
    
    \item $V=\vki$. Then $V_0\in Pa_M(Z_j)\setminus Pa_M(\vki)$.
    Without loss of generality, suppose $\vki$ does not change its position between $M$ and $M''$. If it changes then we can use the same procedure as described in \ref{dog_5}.

    Consider $J$ as the set of variable $V'$ in $M''$ where $N_{V'}$ corresponds to the exogenous noise of a variable in $\{V_0\}\cup (An_{M}(\vki)\cap \setV^C)$, and consider $K = Pa_{M}(\vki)\cap \setV^C$. We have $\ww{\vki}=|K|$. However, any bottleneck from $J$ to $K$ includes one variable between $V'$ and $f(V')$, for all $V'\in K$. Additionally, it needs to cover the edge $V_0\to V$. Therefore the size of the minimal bottleneck must be at least $|K|+1$.
    
    \item $V=\vkl$. Similarly, we assume that $\vki$ does not change its position between $M$ and $M''$. Consider $J$ as the set of variable $V'$ in $M''$ where $N_{V'}$ corresponds to the exogenous noise of a variable in $\{V_0\}\cup (An_{M}(\vkl)\cap \setV^C)$, and consider $K = Pa_{M}(\vkl)\cap \setV^C$. We have $\ww{\vkl}=|K|$. However, any bottleneck from $J$ to $K$ includes one variable between $V'$ and $f(V')$, for all $V'\in K$. Additionally, it needs to cover the edge $V_0\to V$. Therefore the size of the minimal bottleneck must be at least $|K|+1$.
\end{itemize}

\section*{Data availability}
No external dataset was used in this study. The numerical results are based on simulated data generated from the models and settings described in Section~6.

\bibliographystyle{elsarticle-harv}
\bibliography{MyLib}

\end{document}